\documentclass[acmtog]{acmart}
\acmSubmissionID{1200}

\usepackage{booktabs} 
\usepackage{graphicx}
\usepackage{hyperref}
\usepackage[capitalize,nameinlink]{cleveref}
\usepackage{multirow}
\usepackage{colortbl}
\usepackage{xspace}

\definecolor{yellow}{rgb}{1, 1, 0.7}
\definecolor{orange}{rgb}{1, 0.85, 0.7}
\definecolor{tablered}{rgb}{1, 0.7, 0.7}
\definecolor{red}{rgb}{1, 0, 0}

\newcommand{\revise}{\textcolor{black}}
\newcommand{\best}{\cellcolor{tablered}}
\newcommand{\sbest}{\cellcolor{orange}}
\newcommand{\tbest}{\cellcolor{yellow}}

\usepackage[ruled]{algorithm2e} 

\SetAlFnt{\small}
\SetAlCapFnt{\small}
\SetAlCapNameFnt{\small}
\SetAlCapHSkip{0pt}

\acmJournal{TOG}

\begin{document}
\title{Uni3C: Unifying Precisely 3D-Enhanced Camera and Human Motion Controls for Video Generation}

\author{ChenJie Cao}
\orcid{0000-0003-3916-2843}
\affiliation{%
 \institution{Alibaba DAMO Academy, Fudan University, Hupan Lab}
 \country{China}
 }
\email{ccjdurandal422@163.com}
\author{JingKai Zhou}
\affiliation{%
 \institution{Alibaba DAMO Academy, Hupan Lab}
 \country{China}
}
\author{ShiKai Li}
\affiliation{%
 \institution{Alibaba DAMO Academy, Hupan Lab}
 \country{China}
}
\author{JingYun Liang}
\affiliation{%
 \institution{Alibaba DAMO Academy, Hupan Lab}
 \country{China}
}
\author{ChaoHui Yu}
\authornote{Project lead.}
\affiliation{%
 \institution{Alibaba DAMO Academy, Hupan Lab}
 \country{China}
}
\author{Fan Wang}
\affiliation{%
 \institution{Alibaba DAMO Academy}
 \country{China}
}
\author{Yanwei Fu}
\authornote{Corresponding author. Prof. Yanwei Fu is with the Institute of Trustworthy Embodied Al, and the School of Data Science, Fudan University.}
\affiliation{%
 \institution{Fudan University, Shanghai Innovation Institute}
 \country{China}
}
\email{yanweifu@fudan.edu.cn}
\author{XiangYang Xue}
\affiliation{%
 \institution{Fudan University}
 \country{China}
}

\renewcommand\shortauthors{ChenJie, Cao. et al}

\begin{teaserfigure}
\centering
\includegraphics[width=1.0\linewidth]{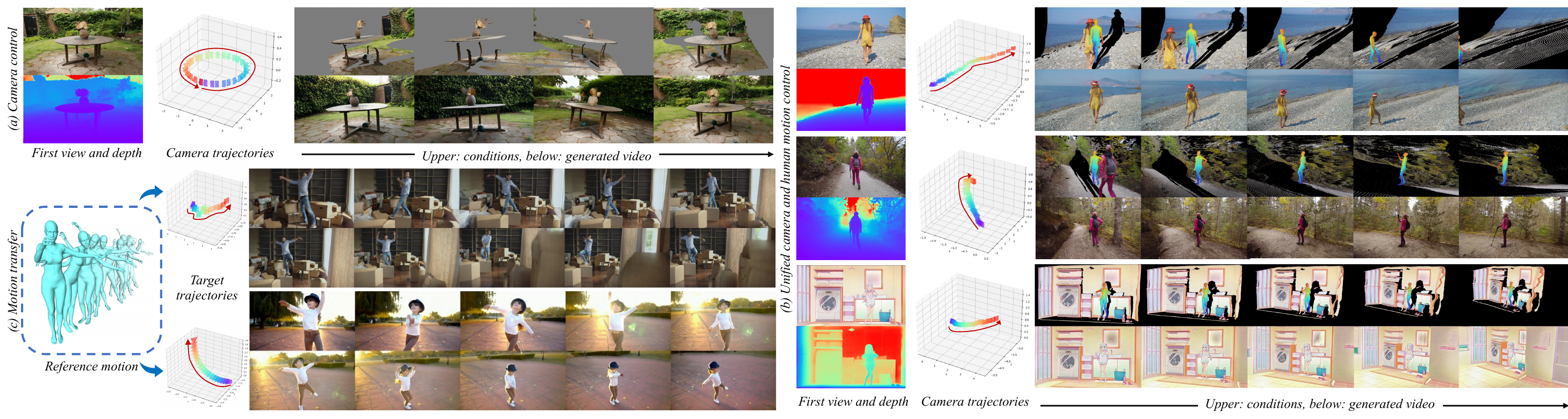}
\vspace{-0.25in}
   \caption{Given a single-view image across various domains (\textit{e.g.}, real-world, text-to-image, animation), we first extract the monocular depth and focal length of it via Depth-Pro~\cite{Bochkovskii2024depthpro} and then achieve point clouds. Then, the proposed Uni3C can generate impressive videos under arbitrary (a) camera trajectories, (b) human motion characters (SMPL-X~\cite{pavlakos2019expressive}), or both of these conditions.
   (c) Uni3C further supports the camera-controlled motion transfer. 
   Please review the videos in the supplementary for more details.
    \label{fig:teaser}}
\end{teaserfigure}

\begin{abstract}
Camera and human motion controls have been extensively studied for video generation, but existing approaches typically address them separately, suffering from limited data with high-quality annotations for both aspects.
To overcome this, we present \textbf{Uni3C}, a unified 3D-enhanced framework for precise control of both camera and human motion in video generation.
Uni3C includes two key contributions. 
First, we propose a plug-and-play control module trained with a frozen video generative backbone, PCDController, which utilizes unprojected point clouds from monocular depth to achieve accurate camera control. 
By leveraging the strong 3D priors of point clouds and the powerful capacities of video foundational models, PCDController shows impressive generalization, performing well regardless of whether the inference backbone is frozen or fine-tuned. 
This flexibility enables different modules of Uni3C to be trained in specific domains, \textit{i.e.}, either camera control or human motion control, reducing the dependency on jointly annotated data.
Second, we propose a jointly aligned 3D world guidance for the inference phase that seamlessly integrates both scenic point clouds and SMPL-X characters to unify the control signals for camera and human motion, respectively.
Extensive experiments confirm that PCDController enjoys strong robustness in driving camera motion for fine-tuned backbones of video generation. 
Uni3C substantially outperforms competitors in both camera controllability and human motion quality. Additionally, we collect tailored validation sets featuring challenging camera movements and human actions to validate the effectiveness of our method.
\revise{Codes are released at \url{https://github.com/alibaba-damo-academy/Uni3C}.}
\end{abstract}

%
%


%
%


\maketitle

\section{Introduction}

Recent advancements in foundational video diffusion models (VDMs)\\
\citep{Blattmann2023svd,yang2025cogvideox,Tim2024sora,kling,runwaygen3,kong2024hunyuanvideo,wang2025wan} have unlocked unprecedented capabilities in creating dynamic and realistic video content.
A significant challenge in this field is achieving controllable video generation, a feature with broad applications in virtual reality, film production, and interactive media.
In this paper, we focus on two aspects of controllable video generation:  camera control~\cite{He2024Cameractrl,yang2024direct,zheng2024cami2v,bahmani2024ac3d,yu2024viewcrafter,ren2025gen3c} and human motion control~\cite{hu2024animate,wang2024humanvid,magicanimate,realisdance,animatex,animateanyone2}—both of which are critical and interdependent in real-world scenarios.

Recent pioneering works have extensively studied controlling camera trajectories for VDMs through explicit conditions like Pl{\"u}cker ray~\cite{He2024Cameractrl,bahmani2024ac3d,zheng2024cami2v,he2025cameractrl,liang2024wonderland}, point clouds~\cite{yu2024viewcrafter,ren2025gen3c,li2025realcam,feng2024i2vcontrol,popov2025camctrl3d}.
Concurrently, controllable human animation also attracted a lot of attention based on poses~\cite{hu2024animate,magicanimate,animatex,animateanyone2} or SMPL characters~\cite{realisdance,champ,zhou2025RealisDance}.
Despite these advancements, several challenges remain: 
1) Most approaches deeply hack the inherent capacities of VDMs, which have been trained with domain-specific data and conditions, inevitably undermining the generalization to handle out-of-distribution scenarios.
2) Very few works investigate the joint control of both camera trajectories and human motions. This requires diverse camera trajectories in human-centric videos with high-quality annotations~\cite{wang2024humanvid}, which are often expensive to obtain.
3) There is a lack of explicit and synchronized guidance that incorporates strong 3D-informed priors to control both camera movements and human motions concurrently. Relying on separate conditions, like point clouds and SMPL, struggles to represent physically reasonable interactions and positional relations between characters and environments, resulting in conflicting guidance and suboptimal outcomes.

To address these challenges, we present \textbf{Uni3C}, a novel framework that \textbf{Uni}fies precise \textbf{3}D-enhanced camera and human motion \textbf{C}ontrols for video generation as shown in \Cref{fig:teaser} via two key innovations.
Firstly, we propose the insight that controlling camera trajectories for powerful foundational VDMs can be achieved by lightweight, trainable modules with \textit{rich informed guidance} and \textit{reasonable training strategies}. By avoiding hacking the underlying capacities of VDMs, our camera-controlling model can be directly generalized to versatile downstream tasks rather than costly joint training and extensive Structure from Motion (SfM) labeling~\cite{schoenberger2016mvs,zhao2022particlesfm,li2024megasam}.
Secondly, we claimed that camera and human motion controls are inherently interdependent. Therefore, we propose to align their conditions into a \textit{global 3D world} during the inference phase, enabling 3D consistent generation across both domains.

Formally, our Uni3C is built upon the foundational VDM—Wan2.1\\~\cite{wang2025wan}.
For the control of camera trajectories, we propose PCDController, a plug-and-play control module with only 0.95B parameters (compared to 14B of Wan2.1) that operates on unprojected 3D point clouds derived from monocular depth estimation~\cite{Bochkovskii2024depthpro}. 
Thanks to the rich geometric priors of point clouds, PCDController is capable of fine-grained camera control, even when trained on constrained multi-view images and videos with a frozen backbone.
Furthermore, PCDController can be compatible with fine-tuned VDM backbones for versatile downstream tasks. 
This surprising factor supports domain-specific training, \textit{i.e.}, camera and human motion modules can be trained independently without jointly annotated data. 
For the global 3D world guidance, we align scenic point clouds (for camera control) and SMPL-X characters~\cite{pavlakos2019expressive} (for human animation) into the same world-coordinate space via the rigid transformation~\cite{umeyama1991least}, while the 2D keypoints~\cite{xu2023vitpose++} bridge the relation of two presentations.
Note that our alignment enables complicated motion transfer, covering disparate characters, positions, and viewpoints as verified in the last row of \Cref{fig:teaser}.

Extensive experiments validate the efficacy of Uni3C. To evaluate the remarkable generalization of PCDController, we collect an out-of-distribution test set across different domains, where each image has four different camera trajectories. 
For the joint controllability, we build a comprehensive test set of in-the-wild human videos. GVHMR~\cite{shen2024world} is used to extract SMPL-X as the condition, while three complex and random camera trajectories are assigned for each video.
VBench++~\cite{huang2024vbench++} is employed to verify the overall performance of our method. Uni3C significantly outperforms other competitors, both quantitatively and qualitatively.

We highlight the key contributions of Uni3C as:
\begin{itemize}
    \item \textbf{PCDController.} A robust, lightweight camera control module is proposed, which enjoys strong 3D priors from point clouds, compatible with both frozen or adapted VDMs.
    \item \textbf{Global 3D World Guidance.} A unified inference framework that aligns scene geometry (point clouds) and human characters (SMPL-X) for 3D-coherent video control.
    \item \textbf{Comprehensive Validation.} We propose new benchmarks and datasets to evaluate challenging camera-human interaction scenarios, demonstrating Uni3C's superiority over existing approaches.
\end{itemize}

\section{Related Work}
\label{sec:related_work}

\subsection{Camera Control for VDMs}
Controlling camera trajectories in video generation has garnered significant attention recently.
Some works focused on injecting camera parameters into VDMs to achieve camera controllability~\cite{wang2024motionctrl,He2024Cameractrl,bahmani2024vd3d,bahmani2024ac3d,zheng2024cami2v,he2025cameractrl,liang2024wonderland}, typically utilizing the Plücker ray presentation.
For instance, VD3D~\cite{bahmani2024vd3d} designed a tailored framework for Diffusion Transformer (DiT)~\cite{peebles2023scalable}, while AC3D~\cite{bahmani2024ac3d} further emphasized the generalization with fewer trainable parameters.
Moreover, DimensionX~\cite{sun2024dimensionx} further decoupled the spatial and temporal control with different LoRAs~\cite{hu2021lora}. 
Despite the progress made by these methods, they cannot control the camera movements precisely, particularly when the case is beyond the training domains with an unknown metric scale.
Thus, other recent works have employed point cloud conditions via training-based~\cite{yu2024viewcrafter,ren2025gen3c,li2025realcam,feng2024i2vcontrol,popov2025camctrl3d} and training-free~\cite{hou2024training,you2024nvs} manners.
However, they fail to accommodate the generalized model design and training strategy to handle imperfect point clouds or out-of-distribution data, especially in motional scenarios involving humans or animals.

\subsection{Unified Control for VDMs}
Recent works have unified multiple conditions to guide video generation~\cite{wang2024motionctrl,feng2024i2vcontrol,geng2024motionprompting,wang2024humanvid,chen2025perception,gu2025diffusion,zheng2025vidcraft3}.
MotionCtrl~\cite{wang2024motionctrl} integrated the camera and object motion controls through separate pose and trajectory injections.
Subsequently, researchers further explored the presentation of conditions, such as point trajectories~\cite{feng2024i2vcontrol}, point tracking~\cite{geng2024motionprompting,gu2025diffusion}, and 3D-aware signals~\cite{chen2025perception}.
Moreover, VidCraft3~\cite{zheng2025vidcraft3} considered the lighting control to VDMs for the first time.
\revise{For the control of both cameras and human motions, some works focus on the behavior transfer based on camera and human motion estimation~\cite{jiang2024cinematic,kocabas2024pace}.}
Humanvid~\cite{wang2024humanvid} first unified them in video generation. 
However, it requires both camera and human guidance from the same source video; otherwise, the human pose would conflict with the camera movements, leading to limited flexibility for the decoupled control.
Despite the promising performance demonstrated by these pioneering approaches, they often rely heavily on joint training under various conditions and well-labeled datasets. 
Furthermore, there has been limited discussion on unified control within foundational VDMs that exceed 10B parameters.
Our work offers a solution to address these issues: unifying existing models for different downstream tasks without the need for costly fine-tuning or fully annotated datasets.
This strategy is particularly well-suited for large VDMs, enabling models to focus more on enhancing performance within their specific domains.

\section{Preliminary: Video Diffusion Models}
\label{sec:perliminary}

We briefly review VDMs and Wan2.1~\cite{wang2025wan} in this section as preliminary knowledge.
VDMs are mainly based on the latent diffusion model~\cite{rombach2022high}, modeling the conditional distribution $p(z_0|c_{txt},c_{img})$, where $z_0$ indicate clean video latent features encoded by 3D-VAE; $c_{txt},c_{img}$ denote the text condition for Text-to-Video (T2V) and the optional image condition for Image-to-Video (I2V), respectively.
The training of VDM involves reversing the diffusion process by the noise estimator $\epsilon_\theta$ as:
\begin{equation}
\label{eq:diffusion}
\min_\theta \mathbb{E}_{z_0,t,\epsilon,c_{txt},c_{img}}[\|\epsilon_\theta(z_t,t,c_{txt},c_{img})-\epsilon\|^2],
\end{equation}
where $\epsilon\sim\mathcal{N}(0,I)$ indicates Gaussian noise; timestep $t\in[0,T_{max}]$; $z_t$ is the intermediate noisy latent state of timestep $t$.
Recently, most VDMs have employed Flow Matching (FM)~\cite{lipman2022flow} as the improved diffusion process with faster convergence and more stable training. 
Based on the ordinary differential equations (ODEs), FM formulates the linear interpolation between $z_0$ and $z_1$, \textit{i.e.}, $z_t=tz_1+(1-t)z_0$, where $t\in[0,1]$ is sampled from the logit-normal distribution.
The velocity prediction $v_\theta$ is written as:
\begin{equation}
\label{eq:flow_matching}
\min_\theta \mathbb{E}_{z_0,t,\epsilon,c_{txt},c_{img}}[\|v_\theta(z_t,t,c_{txt},c_{img})-v_t\|^2],
\end{equation}
where the ground truth velocity denotes $v_t=\frac{dz_t}{dt}=z_1-z_0$.
Additionally, recent foundational VDMs~\cite{Tim2024sora,yang2025cogvideox,kong2024hunyuanvideo,wang2025wan} are built with DiT~\cite{peebles2023scalable} to achieve more capacities for video generation.

\noindent\textbf{Wan2.1}~\cite{wang2025wan} is an open-released VDM with DiT architecture trained with flow matching~\cite{lipman2022flow}.
umT5~\cite{chung2023unimax} is utilized as the multi-language text encoder to inject textual features into Wan2.1 through cross-attention.
For image-to-video, Wan-I2V further incorporates features from CLIP's image encoder~\cite{radford2021learning} to improve the results.
Uni3C is primarily designed for Wan-I2V with 14B parameters, but we empirically find that it is compatible with the Wan-T2V version as verified in \Cref{sec:ablation_and_exp}, showing convincing generalization.

\section{Method}
\label{sec:method}

\begin{figure}
\centering
\includegraphics[width=0.95\linewidth]{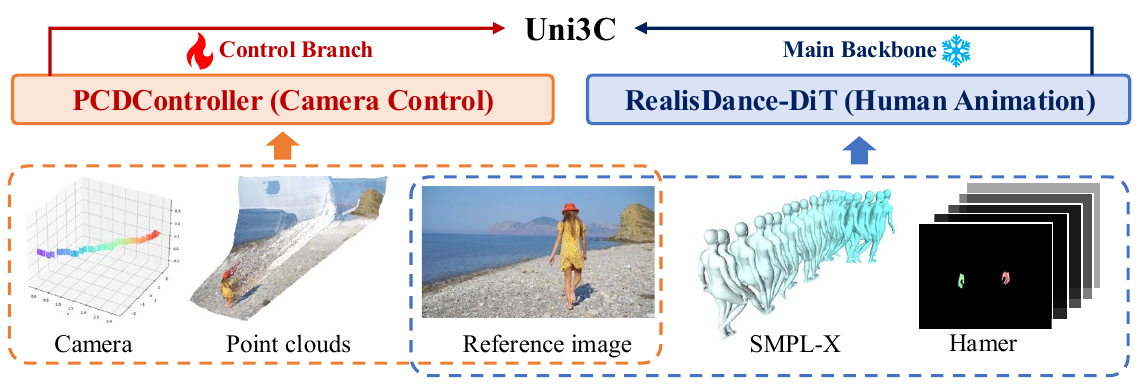}
\vspace{-0.15in}
   \caption{\textbf{The overview of Uni3C,} which adopts multi-modal conditions. The camera, point clouds, and reference image are assigned to the camera control module called PCDController, while the reference image, SMPL-X~\cite{pavlakos2019expressive}, and Hamer~\cite{hamer} are assigned to human animation called RealisDance-DiT~\cite{zhou2025RealisDance}.
    \label{fig:uni3c_conditions}}
\vspace{-0.15in}
\end{figure}

\begin{figure}
\centering
\includegraphics[width=1.0\linewidth]{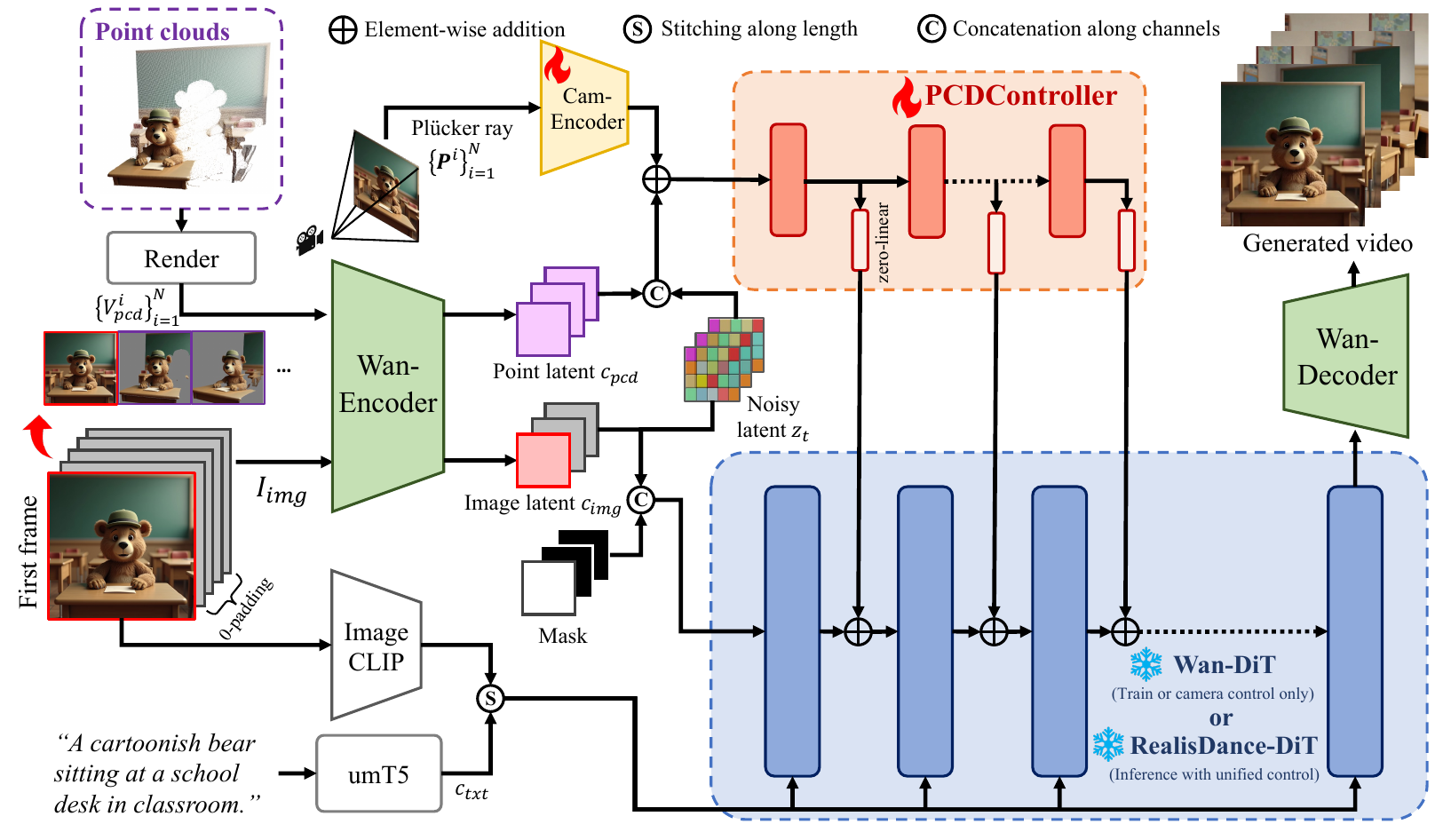}
 \vspace{-0.25in}
   \caption{\textbf{Pipeline of PCDController,} which is built as a lightweight DiT trained from scratch. We first obtain point clouds via monocular depth from the first view. Then, the point clouds are warped and rendered into the video $V_{pcd}$. Input conditions for PCDController comprise rendered $V_{pcd}$, Pl{\"u}cker ray $\mathbf{P}$, and the noisy latent $z_t$. Only the PCDController and camera encoder are trainable in our framework. For inference of unified control over camera and human motions, we directly replace the frozen Wan backbone with RealisDance-DiT~\cite{zhou2025RealisDance} without joint fine-tuning.
    \label{fig:pipeline_pcdcontroller}}
\vspace{-0.15in}
\end{figure}

\paragraph{Overview.}
We first present the overview of Uni3C in \Cref{fig:uni3c_conditions}.
Given a reference view $I_{img}\in\mathbb{R}^{3\times h\times w}$, camera trajectories $\{c_{cam}^i\}_{i=1}^N$ of $N$ target views, and textual condition, Uni3C uses PCDController to produce the target video under specified camera trajectories. This process can be formulated as $p(z_0|c_{txt},c_{img},c_{cam})$, where $z_0$ indicates the clean latent video feature, $c_{txt},c_{img}$ are textual features from umT5 and image latent condition encoded from $I_{img}$, respectively.
We show the pipeline of PCDController in \Cref{fig:pipeline_pcdcontroller} as a core component of Uni3C for generalized camera control (\Cref{sec:pcdcontroller}) across downstream tasks. 
In this work, we focus on human animation (\Cref{sec:human_animation}).
Subsequently, we introduce the global 3D world guidance illustrated in \Cref{fig:world_guidance} to unify both camera and human characters into a consistent 3D space for inference (\Cref{sec:3d_world}).

\subsection{PCDController with 3D Geometric Priors}
\label{sec:pcdcontroller}

\paragraph{Architecture.} 
Following AC3D~\cite{bahmani2024ac3d}, PCDController is designed within a simplified DiT module rather than copying modules and weights from the main backbone as ControlNet~\cite{zhang2023adding}.
To preserve the generalization of Wan-I2V, we follow the insight of \textit{training as few parameters as possible once the effective camera control has been achieved}.
Formally, we reduce the hidden size of PCDController from 5120 to 1024, while zero-initialized linear layers are used to project the hidden size back to 5120 before being added to Wan-I2V.
Moreover, as investigated in~\cite{bahmani2024ac3d,liang2024wonderland}, VDMs mainly determine camera information through shallow layers.
Thus, we only inject camera-controlling features into the first 20 layers of Wan-I2V to further simplify the model. 
Additionally, we discard the textual condition for PCDController to alleviate intractable hallucination and remove all cross-attention modules.
In this way, the overall number of trainable parameters for PCDController is reduced to 0.95B, a significant reduction compared to Wan-I2V (14B).

\paragraph{3D Geometric Priors.}
In contrast to merely utilizing Pl{\"u}cker ray as the camera embedding~\cite{bahmani2024ac3d,liang2024wonderland}, we incorporate much stronger 3D geometric priors to compensate the model simplification, \textit{i.e.}, videos $\{V_{pcd}^i\}_{i=1}^N\in\mathbb{R}^{N\times3\times h\times w}$ rendered from point clouds under given camera trajectories.
Specifically, we first use Depth-Pro~\cite{Bochkovskii2024depthpro} to extract the monocular depth map from the reference view. 
We then align this depth map into a metric representation using SfM annotations~\cite{schoenberger2016mvs} or multi-view stereo~\cite{cao2024mvsformer++}.
Following~\cite{cao2024mvgenmaster}, we employ RANSAC to derive the rescale and shift coefficients, preventing the collapse of constant depth outcomes.
Subsequently, the point clouds $X_{pcd}\in\mathbb{R}^{hw\times3}$ are got by unprojecting all 2D pixels from $I_{img}$ into the world coordinate via its metric depth $\hat{D}_{img}$ as follows:
\begin{equation}
\label{eq:unproject}
X_{pcd}(x)\simeq R_{c\rightarrow w}\hat{D}_{img}(x) K^{-1}\hat{x},
\end{equation}
where $x$ denotes the 2D coordinates of $I_{img}$, while $\hat{x}$ refers to the homogeneous coordinates; $K,R_{c\rightarrow w}$ mean the intrinsic and extrinsic cameras of the reference view, respectively.
After that, we render $\{V_{pcd}^i\}_{i=2}^{N}$ for the remaining ($N-1$) views by PyTorch3D according to their respective camera intrinsics and extrinsics.
Note that the first rendering corresponds to the reference image, \textit{i.e.}, $V_{pcd}^1=I_{img}$, to confirm the identity.
We apply $V_{pcd}$ to the 3D-VAE of Wan2.1 to achieve $c_{pcd}$ as the point latent condition.
To further handle the significant viewpoint changes, which may extend beyond the point clouds' visibility of the first frame, PCDController also includes Pl{\"u}cker ray embedding~\cite{xu2023dmv3d}, $\{\mathbf{P}^i\}_{i=1}^N\in\mathbb{R}^{N\times 6\times h \times w}$, as the auxiliary condition. $\{\mathbf{P}^i\}_{i=1}^N$ is encoded by a small camera encoder, comprising causal convolutions and a 4-8-8 downsampling factor to keep the same sequential length as 3D-VAE outputs.
The distribution modeling of PCDController is  $p(z_0|c_{txt},c_{img},c_{pcd},\mathbf{P})$.

\begin{figure}
\centering
\includegraphics[width=1.0\linewidth]{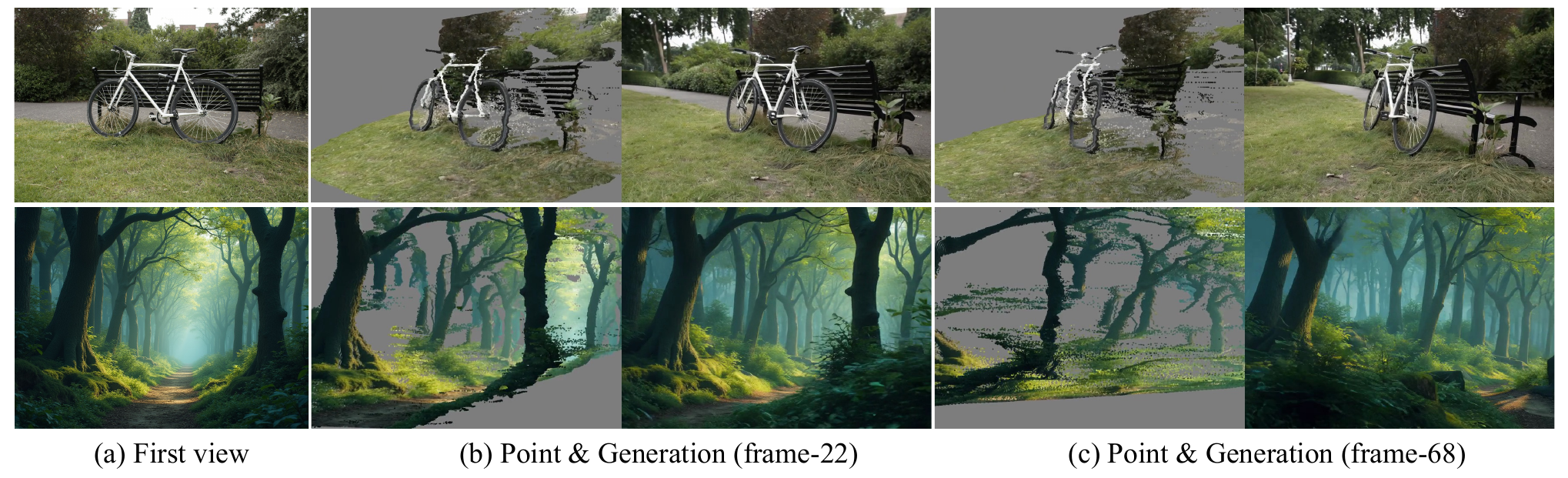}
\vspace{-0.2in}
   \caption{\textbf{Results of PCDController with imperfect point clouds.}
   Benefiting from the well-preserved capacity of VDM, PCDController enjoys robust generation with inferior point clouds. 
    \label{fig:impefect_depth}}
\vspace{-0.2in}
\end{figure}

\paragraph{Discussion.}
We empirically find that our method demonstrates robust performance even when working with imperfect point clouds obtained through monocular depth unprojection. 
In this context, the point clouds serve as the primary camera control signal, facilitating the convergence of training rather than dominating the multi-view geometric and textural generations, as illustrated in \Cref{fig:impefect_depth}.
Moreover, the lightweight PCDController retains its precise camera control capability \emph{over versatile fine-tuned Wan backbones, even without joint training}.
This flexibility allows for a range of downstream applications, showcasing the robustness of our approach.


\subsection{Human Animation}
\label{sec:human_animation}

In this paper, we explore the unified control through two human animation approaches, both of which are built on the Wan2.1 framework, targeting I2V and T2V, respectively. While these methods are not the primary focus of our work, we provide a brief introduction here.
Formally, the concurrently pioneering work, RealisDance-DiT~\cite{zhou2025RealisDance}, directly replaces the Wan-I2V backbone for high-quality human animation during inference.
\revise{RealisDance-DiT incorporates SMPL-X~\cite{pavlakos2019expressive} and Hamer~\cite{hamer} as additional input conditions through newly zero-initialized layers to guide human motions, while camera-control features are added via the external PCDController as shown in \Cref{fig:uni3c_conditions}.}
To ensure the flexibility for motion transfer, RealisDance-DiT randomly selects the reference frame in the video sequence, which is not perfectly aligned with the given SMPL-X. 
RealisDance-DiT only trains self-attention modules and patchify encoders to confirm the generalization.
However, we clarify that integrating the control branch trained within different backbones is still challenging, requiring a generalized control branch like our proposed PCDController.
Furthermore, we tried another version, RealisDance-DiT-T2V, based on Wan-T2V without reference image conditions to explore the generalization of PCDController.
Remarkably, PCDController adapts successfully to Wan-T2V, empowering it with impressive I2V ability.

\subsection{Global 3D World Guidance}
\label{sec:3d_world}

\begin{figure*}
\centering
\includegraphics[width=1.0\linewidth]{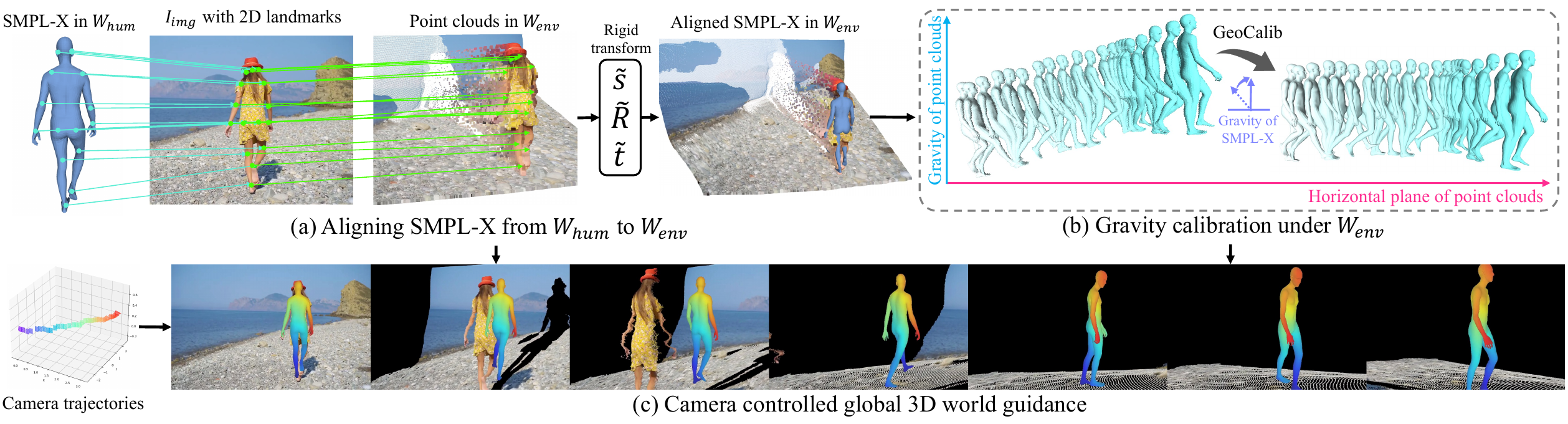}
\vspace{-0.25in}
   \caption{\textbf{Overview of global 3D world guidance.}
   (a) We first align SMPL-X characters from the human world space $W_{hum}$ to the environment world space $W_{env}$ with dense point clouds. (b) GeoCalib~\cite{veicht2024geocalib} is used to calibrate the gravity direction of SMPL-X. (c) Rigid transformation coefficients $\tilde{s},\tilde{R},\tilde{t}$ are employed to align the whole SMPL-X sequence. We re-render all aligned conditions under specific camera trajectories as the global 3D world guidance.
    \label{fig:world_guidance}}
\vspace{-0.125in}
\end{figure*}

\paragraph{Definition.}
As illustrated in \Cref{fig:uni3c_conditions}, Uni3C adopts multi-modal conditions, including camera, point clouds, reference image, SMPL-X~\cite{pavlakos2019expressive}, and Hamer~\cite{hamer}. The first three conditions are used for PCDController, while the latter three are used for human animations.
We employ GVHMR~\cite{shen2024world} to recover SMPL-X characters. One can also retrieve desired motion sequences from motion datasets~\cite{plappert2016kit,punnakkal2021babel,guo2022generating,lin2023motion} or generate new motions through text-to-motion models~\cite{jiang2023motiongpt,guo2024momask,barquero2024seamless}.
Although most of the conditions above are formulated as 3D presentations, they stay in two different world coordinates. 
We define the point cloud world coordinate as the \textbf{environmental world space} $W_{env}$, which is the main world space controlled by cameras, while the SMPL-X is placed in the \textbf{human world space} $W_{hum}$.
It is non-trivial to control the camera across two different world spaces consistently.
For example, determining the initial camera position within $W_{hum}$ is particularly ambiguous, especially for tasks involving motion transfer.
Moreover, to facilitate flexible control over separate camera and human motions without conflicts, human conditions should be re-rendered under new camera trajectories.
Therefore, we propose the global 3D world guidance that places the human condition into the ``environment'', \textit{i.e.}, aligning SMPL-X from $W_{hum}$ to $W_{env}$ as shown in \Cref{fig:world_guidance}(b).

\paragraph{Multi-Modal Alignment.}
Fortunately, 2D human pose keypoints subtly bridge $W_{hum}$ and $W_{env}$.
Formally, we first estimate 17 human keypoints $\{\mathbf{k}^{i}_{2D}\}_{i=1}^{17}\in\mathbb{R}^{17\times 2}$ from the reference view $I_{img}$ by ViTPose++~\cite{xu2023vitpose++}. 
Then, we unproject 2D keypoints into $W_{env}$ to obtain 3D keypoints $\{\mathbf{k}^{i}_{env}\}_{i=1}^{17}\in\mathbb{R}^{17\times 3}$ through metric monocular depth $\hat{D}_{img}$ and the intrinsic camera of the reference image.
For SMPL-X in $W_{hum}$, the COCO17 regressor~\cite{shen2024world} is utilized to project the first frame's SMPL-X character into $\{\mathbf{k}^{i}_{hum}\}_{i=1}^{17}$ corresponding to the same human keypoints.
Consequently, a least-squares estimation~\cite{umeyama1991least} based rigid transformation can be used to align $\mathbf{k}_{hum}$ to $\mathbf{k}_{env}$ as follows:
\begin{equation}
\label{eq:rigid_trans}
\min_{\tilde{s},\tilde{R},\tilde{t}}\sum_{i=1}^{17}w_i\|(\tilde{s}\tilde{R}(\mathbf{k}^i_{hum})^T+\tilde{t})^T-\mathbf{k}^i_{env}\|^2,
\end{equation}
where $\tilde{s},\tilde{R},\tilde{t}$ indicate the optimized scaling factor, rotation matrix, and translation vector, respectively. $w_i$ denotes the confidence weight of 2D keypoint $\mathbf{k}^i_{2D}$. We discard any keypoints with confidence below 0.7, as they typically degrade alignment quality.
Once the transformation parameters $\tilde{s},\tilde{R},\tilde{t}$ are determined, we apply them to all other SMPL-X sequences under the assumption that they share the same rigid transformation.
However, even minor orientation errors can accumulate, leading to physically unrealistic motion trajectories, such as ascending into the sky or descending into the ground.
To address this, we adopt GeoCalib~\cite{veicht2024geocalib} to estimate the gravity direction in $W_{env}$, which is then employed to calibrate the SMPL-X to ensure parallel gravity directions, as illustrated in \Cref{fig:world_guidance}(b).
For the alignment of Hamer~\cite{hamer}, which shares common vertices with the hand parts of SMPL-X, Hamer can also be aligned to $W_{env}$ through the rigid transformation (\Cref{eq:rigid_trans}). Additionally, we address the issue of hand occlusion for Hamer by masking occluded hand parts based on the rendered depth from SMPL-X.
After the alignments of SMPL-X and Hamer sequences, we place all conditions into $W_{env}$, establishing the global 3D world guidance that allows for rendering concurrently controlled conditions under arbitrary camera trajectories and human motions as shown in \Cref{fig:world_guidance}(c).
Finally, these re-rendered conditions are sent to PCDController and RealisDance-DiT for generated outcomes, as shown in~\Cref{fig:uni3c_conditions}.

\section{Experiments}
\label{sec:experiments}

\subsection{Implementation Details}

PCDController is trained with the frozen Wan-I2V~\cite{wang2025wan} on multi-resolution images scaled from [$480\times768$, $512\times720$, $608\times608$, $720\times512$, $768\times480$] of 81 frames. The learning rate is warmed up to 1e-5 for 400 steps and then fixed. We train the model for 6,000 steps with batch size 32, while more training steps would slightly hurt Wan-I2V's generalization. The training is accomplished with 64 H100 GPUs for 40 hours.
We also provided results based on CogVideoX-5B-I2V~\cite{yang2025cogvideox}, training for 20k steps with batch size 16.
During the training, we randomly drop 10\% texts, as well as 5\% point cloud renderings and Pl{\"u}cker embeddings.
For inference, we set the classifier-free guidance scale to 5.0 for textual conditions, keeping other guidance on the default scale 1. 

\paragraph{Datasets.}
\label{sec:datasets}
To ensure the generalization of PCDController, we collect large-scale training data for camera control, including DL3DV\\~\cite{ling2024dl3dv}, RE10K~\cite{zhou2018Stereo}, ACID~\cite{liu2021infinite}, Co3Dv2~\cite{reizenstein21co3d}, Tartainair~\cite{tartanair2020iros}, Map-Free-Reloc~\cite{arnold2022map}, WildRGBD~\cite{xia2024rgbd}, COP3D~\cite{sinha2023common}, UCo3D~\cite{liu2025uncommon}. This comprehensive dataset encompasses various scenarios, featuring static and dynamic scenes, as well as object-level and scene-level environments. Furthermore, all datasets are annotated with metric-aligned monocular depth through the way proposed in~\cite{cao2024mvgenmaster} or are provided with ground-truth depth.

\subsection{Results of Camera Control}
\label{sec:exp_cam}

\begin{table*}
\centering
\caption{\textbf{Quantitative results of camera control.} VBench++ scores (\%) are normalized (higher is better). Injected camera features are divided as Pl{\"u}cker ray and point clouds (Pcd). $\dagger$ denotes the results with challenging 360$^\circ$ camera rotations. Results of MotionCtrl and CameraCtrl are tested with the I2V version re-trained by~\cite{zheng2024cami2v}.
Methods in \textbf{bold} are the final setting of PCDController. 
\label{tab:cam_exp1}}
\vspace{-0.1in}
\small
\setlength{\tabcolsep}{2pt}
\begin{tabular}{l|cc|ccccccccc|ccc}
\toprule 
 & \multicolumn{2}{c|}{Camera} & Overall & Subject & Bg & Aesthetic & Imaging & Temporal & Motion & I2V & I2V & \multirow{2}{*}{ATE$\downarrow$} & \multirow{2}{*}{RPE$\downarrow$} & \multirow{2}{*}{RRE$\downarrow$}\tabularnewline
 & Plücker & Pcd & Score & Consist & Consist & Quality & Quality & Flicker & Smooth & Subject & Bg &  &  & \tabularnewline
\midrule 
MotionCtrl~\scriptsize\cite{wang2024motionctrl} &  &  & 82.48 & 89.42 & 91.64 & 54.99 & 55.02 & 91.93 & 95.85 & 89.92 & 91.11 & 0.345 & 0.263 & 2.547\tabularnewline
CameraCtrl~\scriptsize\cite{He2024Cameractrl} & $\checkmark$ &  & 83.69 & 90.17 & 92.21 & 55.71 & 54.13 & 93.32 & 96.65 & 93.34 & 94.01 & 0.354 & 0.239 & 2.306\tabularnewline
CamI2V~\scriptsize\cite{zheng2024cami2v} & $\checkmark$ &  & 86.52 & 91.45 & 92.65 & \best 60.91 & 66.22 & 93.34 & 97.01 & 95.17 & 95.45 & 0.322 & 0.247 & 1.653\tabularnewline
ViewCrafter~\scriptsize\cite{yu2024viewcrafter} &  & $\checkmark$ & 85.39 & 89.69 & 91.68 & 55.13 & 64.33 & 92.94 & 97.66 & 95.59 & 96.11 & 0.210 & 0.117 & 0.873\tabularnewline
SEVA~\scriptsize\cite{zhou2025stable} & $\checkmark$ &  & 87.39 & 91.86 & \tbest 93.36 & 56.79 & 68.43 & \sbest 95.74 & \sbest 98.59 & 97.08 & 97.23 & \best{0.077} & \sbest{0.029} & \sbest{0.223}\tabularnewline
Ours (CogVideoX) & $\checkmark$ &  & 86.48 & 93.17 & \sbest 93.02 & 55.00 & 66.75 & 95.10 & 98.36 & 94.45 & 95.94 & 0.356 & 0.162 & 1.280\tabularnewline
Ours (CogVideoX) &  & $\checkmark$ & 87.22 & 91.26 & 92.44 & 56.90 & 69.53 & 94.79 & 98.47 & 96.60 & 97.79 & 0.123 & 0.045 & 0.346\tabularnewline
Ours (Wan-I2V) & $\checkmark$ &  & \best 89.16 & \best 94.71 & \best 94.93 & \sbest 60.42 & \best 72.20 & \best 96.51 & 98.51 & \best 97.74 & \best 98.29 & 0.402 & 0.095 & 0.728\tabularnewline
Ours (Wan-I2V) &  & $\checkmark$ & \tbest 87.95 & 91.71 & 92.97 & 58.52 & \tbest 71.12 & 95.51 & \tbest 98.55 & \tbest 97.24 & \tbest 97.96 & \sbest{0.091} & \best{0.028} & \best{0.211}\tabularnewline
\textbf{Ours (Wan-I2V)} & $\checkmark$ & $\checkmark$ & \sbest{88.27} & \tbest 92.20 & \sbest 93.37 & \tbest 58.99 & \sbest 71.96 & \tbest 95.56 & \best 98.66 & \sbest 97.38 & \sbest 98.01 & \tbest 0.102 & \tbest 0.031 & \tbest 0.246\tabularnewline
\midrule
Ours (Wan-I2V)$\dagger$ &  & $\checkmark$ & \sbest 82.70 & \sbest 82.16 & \sbest 88.72 & \best 53.67 & \best 66.95 & \best 91.41 & \best 95.45 & \sbest 90.80 & \sbest 92.47 & \sbest 1.327 & \sbest 0.551 & \sbest 6.334\tabularnewline
\textbf{Ours (Wan-I2V)}$\dagger$ & $\checkmark$ & $\checkmark$ & \best 82.82 & \best 82.31 & \best 88.75 & \sbest 53.56 & \sbest 66.25 & \sbest 90.81 & \sbest 95.23 & \best 92.12 & \best 93.49 & \best{1.010} & \best{0.416} & \best{4.428}\tabularnewline
\bottomrule 
\end{tabular}
\vspace{-0.1in}
\end{table*}

\begin{figure}
\centering
\includegraphics[width=1.0\linewidth]{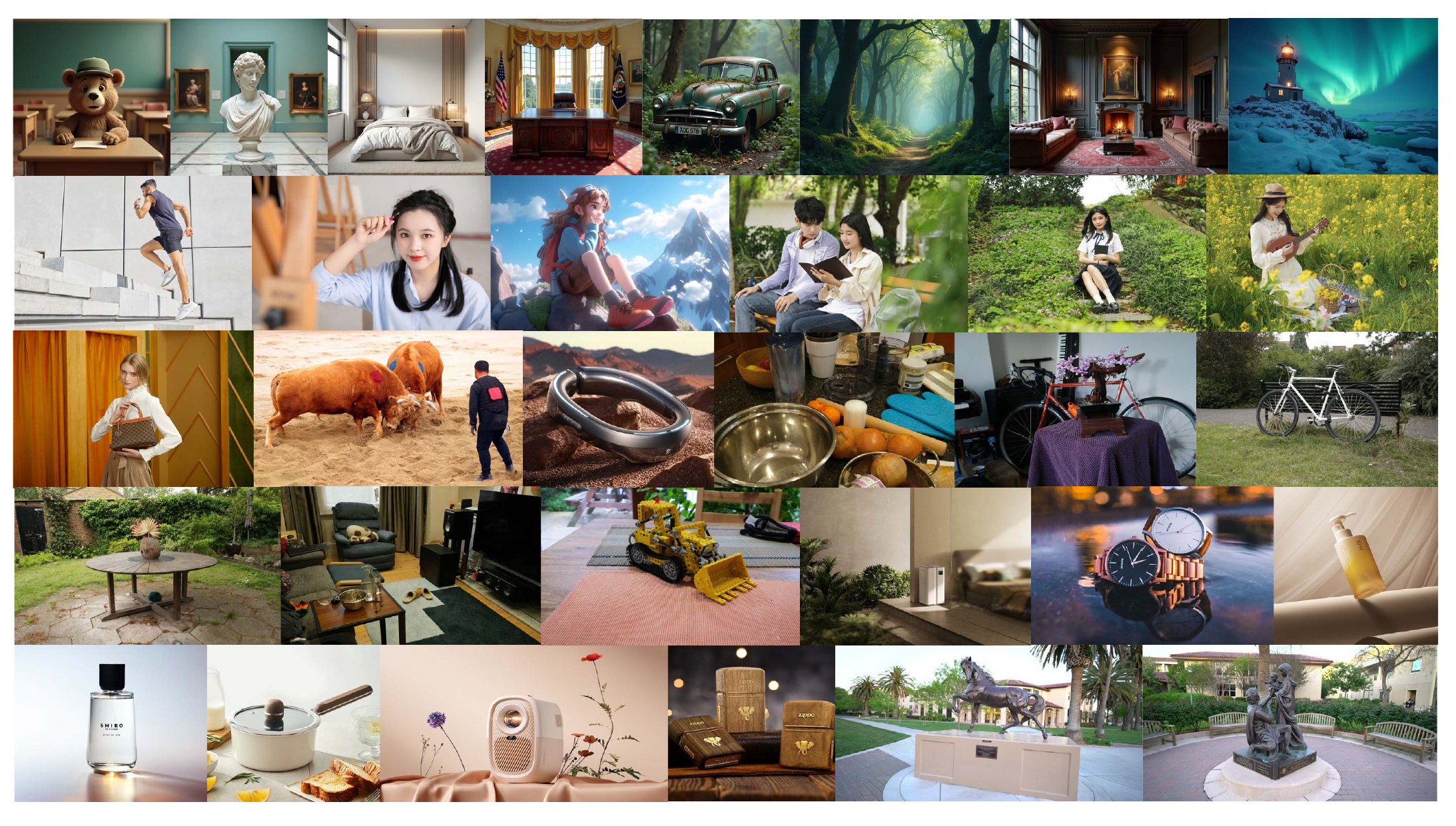}
\vspace{-0.15in}
   \caption{\textbf{Our out-of-distribution benchmark for camera control.} The validation set includes generative, human, scene-level, and object-level images with diverse aspect ratios.
    \label{fig:camera_ood_set}}
\vspace{-0.1in}
\end{figure}

\paragraph{Benchmark.} To evaluate camera control ability, we built an out-of-distribution benchmark with 32 images across various domains, including text-to-image generation\footnote{\url{https://github.com/black-forest-labs/flux}}, real-world~\cite{knapitsch2017tanks,barron2022mip}, object-centric, and human scenes as in \Cref{fig:camera_ood_set}. Each image has four distinct camera trajectories, resulting in 128 test samples. 
\revise{Moreover, we collect a larger benchmark of 500 samples, comprising 50 scene images from megascenes~\cite{tung2024megascenes} and 50 human scenes collected from the Internet. For each image, 5 random trajectories are assigned.}
We used VBench++~\cite{huang2024vbench++} to evaluate the video quality, while absolute translation error (ATE), relative translation error (RPE), and relative rotation error (RRE) are used to verify the camera precision. We utilize VGGT~\cite{wang2025vggt} to produce extrinsic cameras for the generated images, which are then evaluated against the predefined cameras after trajectory alignment.

\begin{figure}[h!]
\centering
\includegraphics[width=0.975\linewidth]{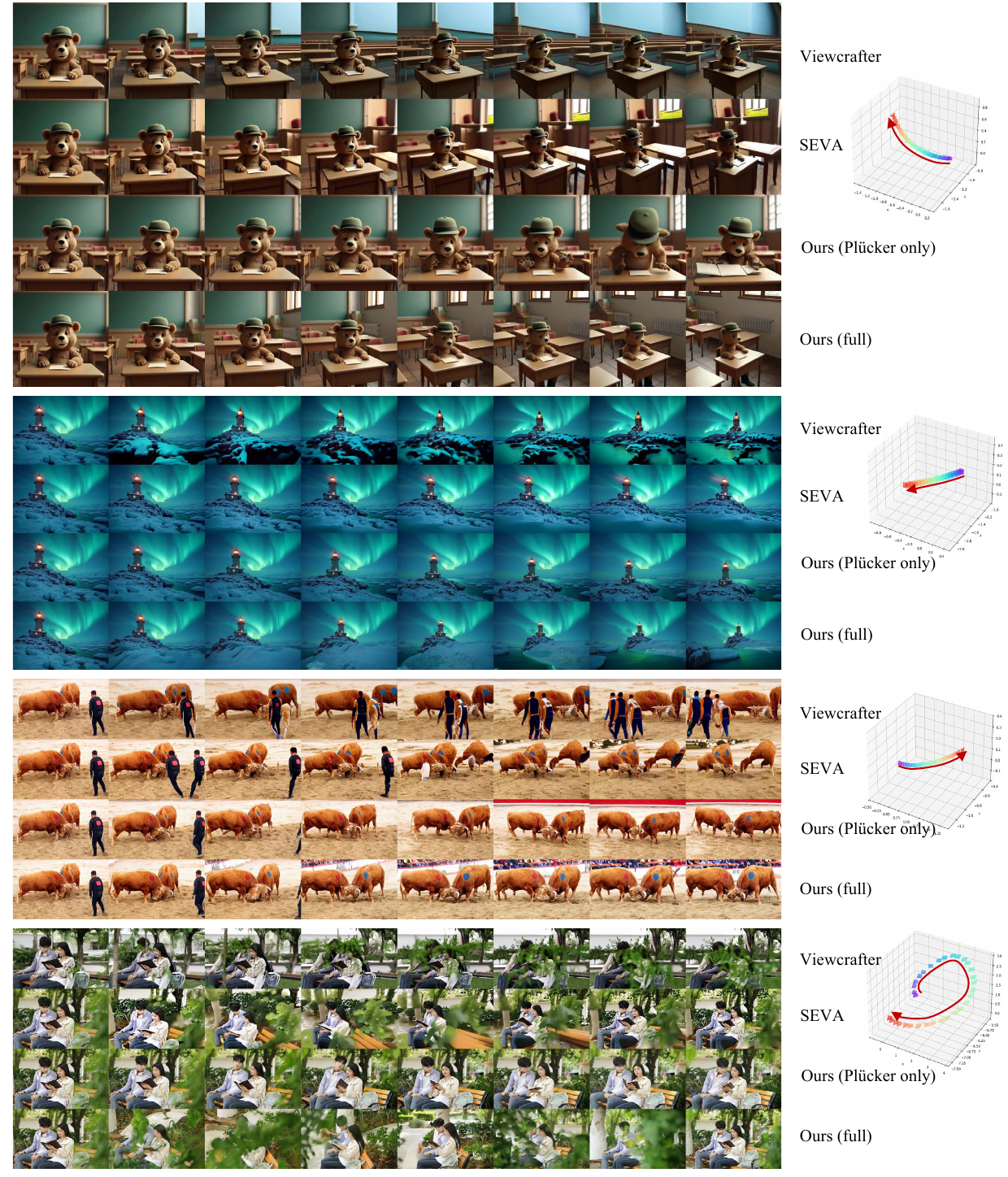}
\vspace{-0.15in}
   \caption{\textbf{Qualitative results of camera control on our benchmark.}
   We compare the proposed PCDController to ViewCrafter~\cite{yu2024viewcrafter}, SEVA~\cite{zhou2025stable}, and our model without point cloud guidance. The leftmost image is the reference condition. ``full'' indicates using both Pl{\"u}cker ray and point clouds as conditions.
    \label{fig:cam_control_results}}
\vspace{-0.15in}
\end{figure}

\begin{figure}[h!]
\centering
\includegraphics[width=0.96\linewidth]{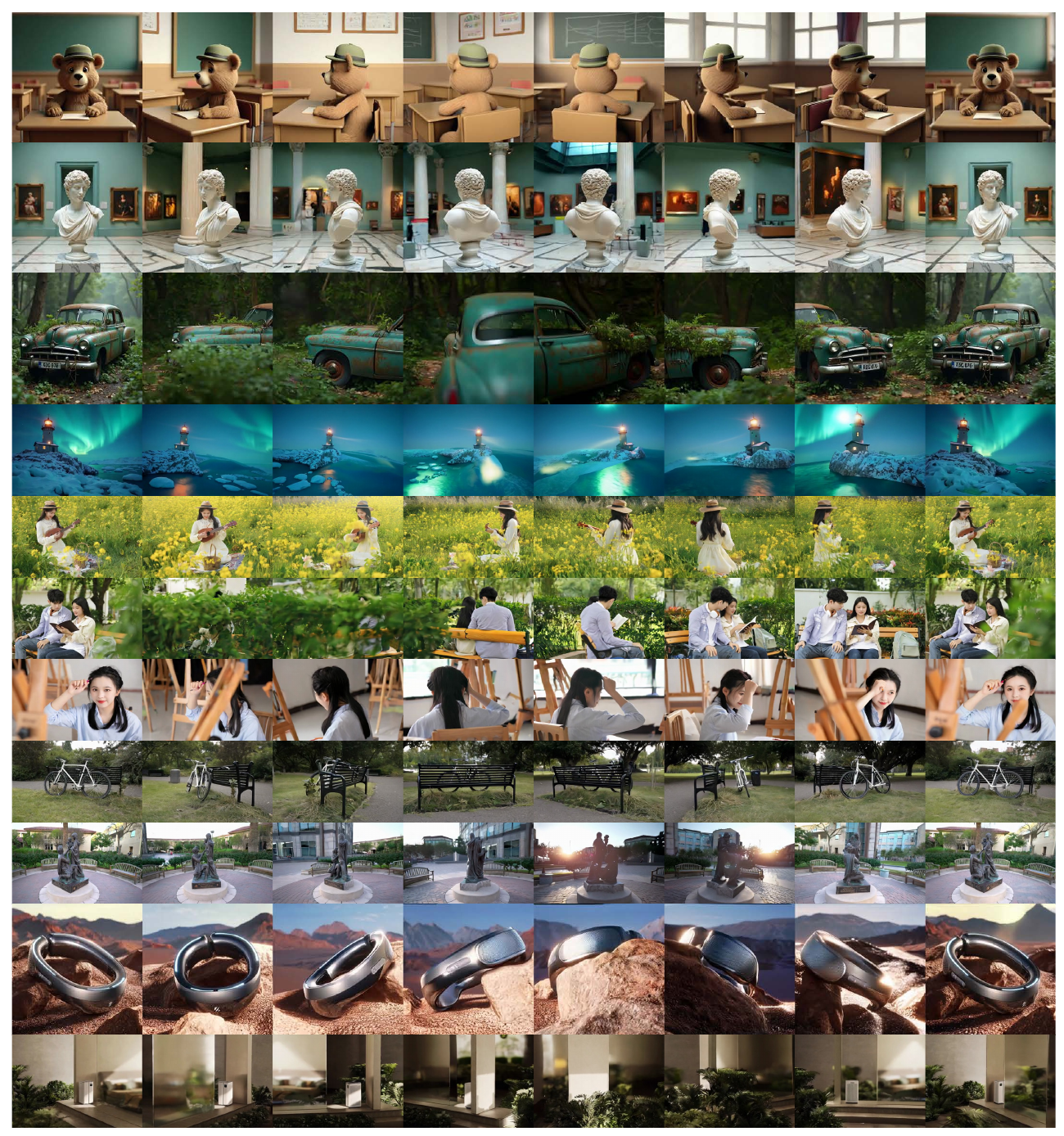}
\vspace{-0.15in}
   \caption{\textbf{Results of the challenging orbital 360$^\circ$ rotations from PCDController.} The leftmost images are the reference views.
    \label{fig:orbit360}}
\vspace{-0.15in}
\end{figure}

\paragraph{Analysis.} We present the quantitative results in \Cref{tab:cam_exp1}, comparing our model with ViewCrafter~\cite{yu2024viewcrafter}, SEVA~\cite{zhou2025stable}, and the CogVideoX~\cite{yang2025cogvideox} version of our framework. The qualitative outcomes are shown in \Cref{fig:cam_control_results}.
Our experiments demonstrate that point clouds significantly enhance the controllability of both the Wan2.1 and CogVideoX, as verified by the improvement of ATE, RPE, and RRE.
Although SEVA achieves precise camera trajectories, it requires massive training with static multi-view data (0.8M iterations), struggling to handle dynamic out-of-distribution scenarios, such as humans and animals, as illustrated in \Cref{fig:cam_control_results}.
We should clarify that our baseline, Wan-I2V with only Plücker ray, suffers from inferior camera movements. 
While this setting achieves a strong VBench overall score, it compromises with poor camera metrics.
Overall, the proposed PCDController achieves the optimal balance between video quality and camera precision. By integrating both Plücker rays and point clouds, it further enhances the performance in challenging scenes featuring substantial viewpoint changes, as validated in \Cref{tab:cam_exp1} and \Cref{fig:orbit360}.
\revise{Furthermore, our method also performs well in the large benchmark on megascene and Internet human images as listed in \Cref{tab:cam_exp2}.}

\begin{table}
\centering
\caption{\revise{\textbf{Camera control results on large benchmark with 500 samples} collected from megascenes~\cite{tung2024megascenes} and Internet human scenes. `Overall' means the average score across all metrics of VBench++. 
\label{tab:cam_exp2}}}
\vspace{-0.15in}
\small
\begin{tabular}{lcccc}
\toprule
 & Overall$\uparrow$ & ATE$\downarrow$ & RPE$\downarrow$ & RRE$\downarrow$\tabularnewline
\midrule
CamI2V~\scriptsize\cite{zheng2024cami2v} & 83.62 & 0.8081 & 0.6858 & 2.5803\tabularnewline
ViewCrafter~\scriptsize\cite{yu2024viewcrafter} & 84.26 & 0.6224 & 0.4098 & 1.2642\tabularnewline
SEVA~\scriptsize\cite{zhou2025stable} & \underline{84.83} & \textbf{0.2473} & \underline{0.0832} & \underline{0.5329}\tabularnewline
PCDController & \textbf{86.62} & \underline{0.2613} & \textbf{0.0828} & \textbf{0.4612}\tabularnewline
\bottomrule
\end{tabular}
\vspace{-0.15in}
\end{table}

\subsection{Results of Unified Camera and Human Motion Control}
\label{sec:exp_unified}

\begin{table*}
\centering
\caption{\textbf{Quantitative results of unified camera and human motion controls.} 
``Aligned'' control means re-rendering human conditions under new camera trajectories in the environmental world space.
$\dagger$ denotes masking out the foreground point clouds of humans.
Method in \textbf{bold} is the final setting of Uni3C.
\label{tab:uni_exp1}}
\vspace{-0.1in}
\small
\setlength{\tabcolsep}{2.5pt}
\begin{tabular}{l|ccc|ccccccc|ccc}
\toprule 
 & \multicolumn{3}{c|}{Control} & Overall & Subject & Bg & Aesthetic & Imaging & Temporal & Motion & \multirow{2}{*}{ATE$\downarrow$} & \multirow{2}{*}{RPE$\downarrow$} & \multirow{2}{*}{RRE$\downarrow$}\tabularnewline
 & Camera & Human & Aligned & Score & Consist & Consist & Quality & Quality & Flicker & Smooth &  &  & \tabularnewline
\midrule 
CamAnimate~\scriptsize\cite{wang2024humanvid} & $\checkmark$ & $\checkmark$ &  & 82.45 & \tbest 89.20 & 90.52 & \tbest 57.42 & 67.40 & 93.94 & 96.20 & 0.619 & 0.419 & 2.035\tabularnewline
RealisDance-DiT~\scriptsize\cite{zhou2025RealisDance} &  & $\checkmark$ & $\checkmark$ & \best{85.21} & \best 93.03 & \best 95.34 & \best 57.89 & 68.71 & \best 97.44 & \best 98.82 & 0.549 & 0.195 & 0.547\tabularnewline
PCDController & $\checkmark$ & & & 83.19 & 89.08 & 91.63 & 57.23 & \tbest 68.27 & 95.22 & 97.71 & \sbest{0.256} & \sbest{0.092} & 0.661\tabularnewline
Uni3C (T2V)$\text{\ensuremath{\dagger}}$ & $\checkmark$ & $\checkmark$ & $\checkmark$ & \tbest 83.34 & 88.45 & 91.45 & \sbest 57.45 & \best 69.84 & 95.21 & 97.63 & 0.296 & 0.098 & 1.167\tabularnewline
Uni3C (T2V) & $\checkmark$ & $\checkmark$ & $\checkmark$ & 83.16 & 88.67 & 91.38 & 56.79 & \sbest 69.42 & 95.14 & 97.57 & \tbest 0.262 & \best{0.083} & \sbest{0.606}\tabularnewline
Uni3C (I2V, unaligned) & $\checkmark$ & $\checkmark$ &  & 82.98 & 89.16 & \tbest 92.40 & 56.02 & 66.78 & \tbest 95.65 & \tbest 97.84 & 0.270 & 0.111 & \tbest 0.639\tabularnewline
\textbf{Uni3C (I2V)} & $\checkmark$ & $\checkmark$ & $\checkmark$ & \sbest{83.43} & \sbest 89.45 & \sbest 93.05 & 57.25 & 67.28 & \sbest 95.70 & \sbest 97.86 & \best{0.251} & \tbest 0.093 & \best{0.490}\tabularnewline
\bottomrule 
\end{tabular}
\end{table*}

\begin{figure*}
\centering
\includegraphics[width=1.0\linewidth]{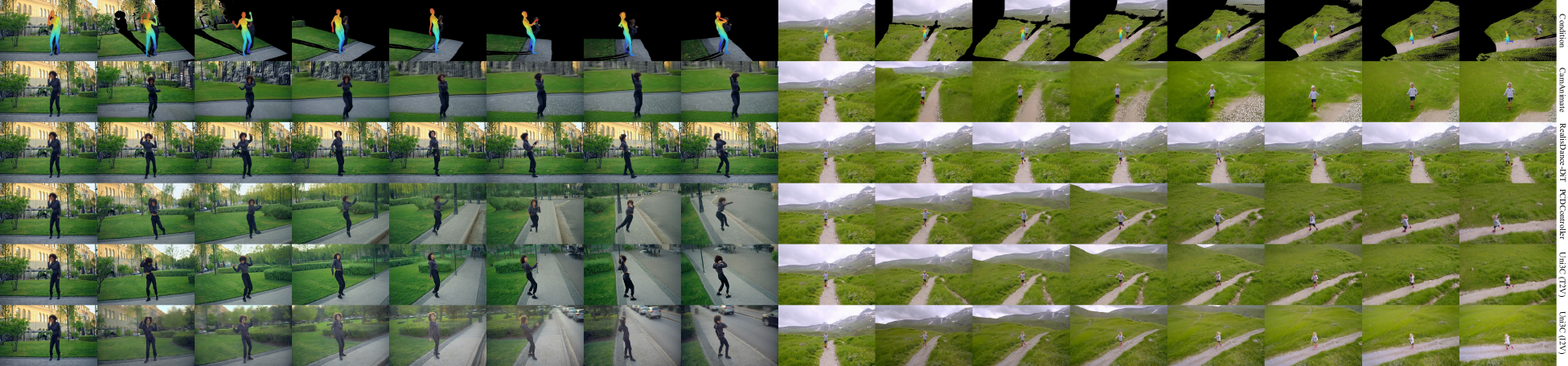}
\vspace{-0.2in}
   \caption{\textbf{Results of unified camera and human motion controls.} Leftmost images are reference views; the first row indicates aligned 3D world guidance.\label{fig:uni_control_results_main}}
\vspace{-0.1in}
\end{figure*}

\paragraph{Benchmark.} We have developed a new benchmark of 50 videos of 720p, each featuring a person performing challenging motions. For guidance, SMPL-X is extracted using GVHMR~\cite{shen2024world}. Each video is assigned three different types of camera trajectories, resulting in a total of 150 test cases. 
To ensure that the person remains within the camera's viewpoint, we employ a follow shooting technique for all test cases, adjusting the camera's position based on the movement of the SMPL-X center. These noisy and subtle movements further increase the difficulty of camera control.
We follow the same camera metrics as mentioned in \Cref{sec:exp_cam}.
To facilitate generalization for motion transfer, RealisDance-DiT is not specifically designed to perfectly recover the first frame aligned with the reference image. Consequently, we remove the metrics that heavily depend on the reference view (I2V subject and background) in \Cref{tab:uni_exp1} for fairness. 
Note that our method performs well in 720p generation, even though it was trained using 480p data.

\begin{figure}[h!]
\centering
\includegraphics[width=0.925\linewidth]{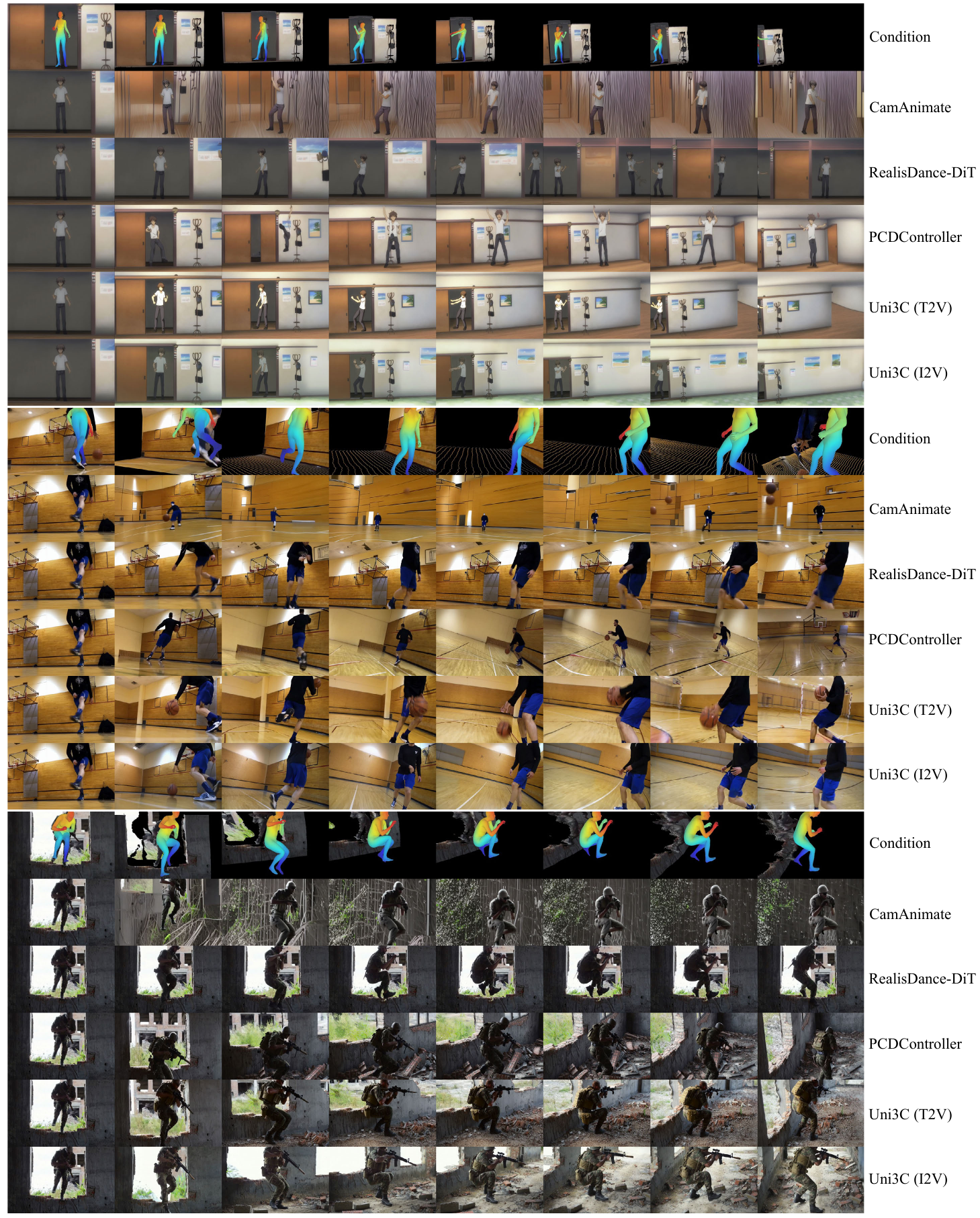}
\vspace{-0.15in}
   \caption{\textbf{More results of unified camera and human motion controls.} Leftmost images are reference views; the first row indicates aligned 3D world guidance.
    \label{fig:union_control_results}}
\vspace{-0.1in}
\end{figure}

\begin{figure}[h!]
\centering
\includegraphics[width=0.9\linewidth]{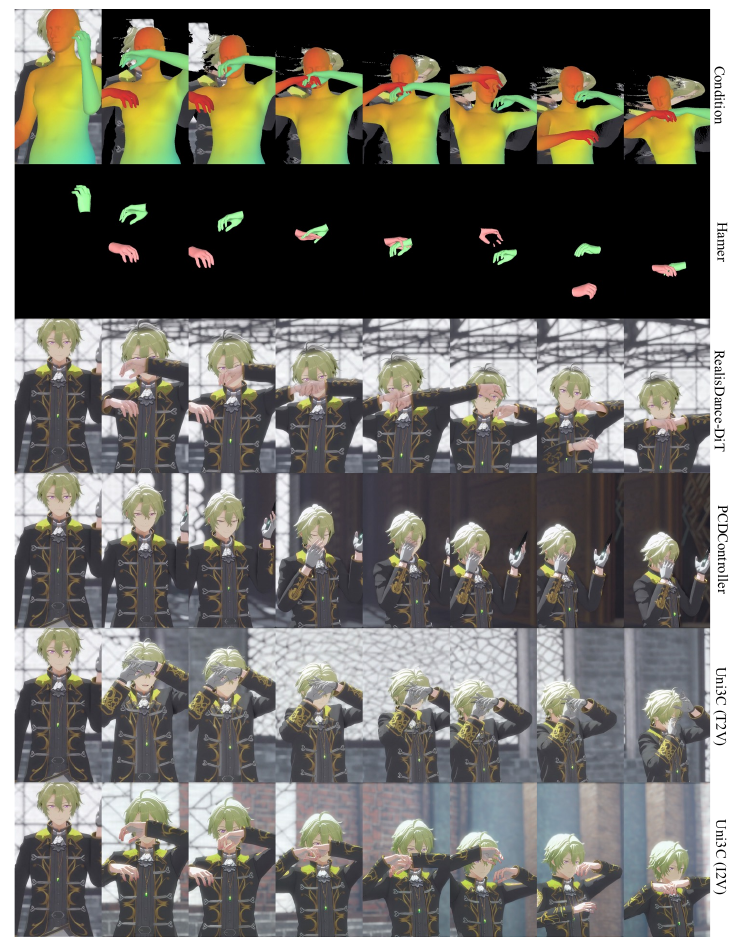}
\vspace{-0.2in}
   \caption{\textbf{Results of unified camera, human motion, and Hamer controls.} The leftmost images are the reference views, while the first and second rows indicate the aligned 3D world guidance and Hamer rendering.
    \label{fig:hand_control_results}}
\vspace{-0.1in}
\end{figure}

\paragraph{Analysis.}
We show quantitative results in \Cref{tab:uni_exp1}, while qualitative results are displayed in \Cref{fig:uni_control_results_main} and \Cref{fig:union_control_results}.
To our knowledge, there are currently very few available models that effectively address the challenge of unified camera and human motion controls. 
We begin by comparing to CamAnimate~\cite{wang2024humanvid}, which enables simultaneous control of both conditions. 
However,  due to the misalignment of body poses extracted relative to predefined camera trajectories, the results of CamAnimate suffer from conflicting human motions and backgrounds.
Moreover, we evaluate Uni3C against the baseline of RealisDance-DiT~\cite{zhou2025RealisDance} and various ablation versions of our model.
Notably, while RealisDance-DiT, which focuses solely on human control, achieves the best visual quality, it struggles to produce accurate camera trajectories, resulting in poor camera metrics. In contrast, Uni3C shows good VBench scores alongside impressive camera metrics.
Note that the proposed PCDController can also be generalized to the T2V model with comparable quality, featuring the robust generalization of PCDController.
Moreover, an interesting insight is revealed from \Cref{fig:union_control_results} and \Cref{tab:uni_exp1} that the aligned SMPL-X characters can further strengthen the camera controllability, while the PCDController trained with I2V formulation also enhances the visual quality and consistency of RealisDance-DiT.
This illustrates the complementary features of these two components.
Additionally, we show that the proposed Uni3C enables control of detailed hand motions under various camera trajectories as in \Cref{fig:hand_control_results}.

\subsection{Ablation Study and Exploratory Discussions}
\label{sec:ablation_and_exp}

\begin{figure}
\centering
\includegraphics[width=1.0\linewidth]{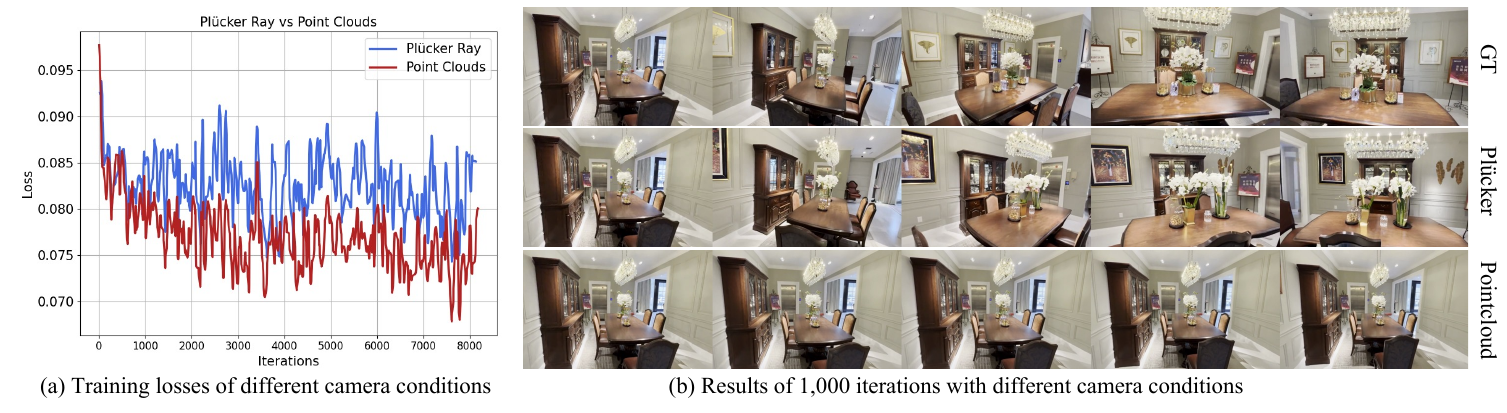}
\vspace{-0.15in}
   \caption{\textbf{Ablation results of Pl{\"u}cker ray and point clouds during training phase.} Point clouds enjoy highly accurate camera control against Pl{\"u}cker ray. \label{fig:pcd_ablation}}
\vspace{-0.1in}
\end{figure}

\paragraph{Pl{\"u}cker Ray vs Point Clouds.}
As shown in \Cref{tab:cam_exp1}, point clouds enjoy significantly more precise camera trajectories. We further clarify that video camera control trained with point clouds achieves much faster convergence with lower training loss, as illustrated in \Cref{fig:pcd_ablation}. Only 1,000 training iterations can hold the general camera trajectory. We empirically find that timing large-scale VDM like Wan-I2V through Pl{\"u}cker ray is difficult. Maybe enabling more trainable parameters would improve the performance, potentially hindering the generalization, which is not considered in this work.
Moreover, as verified in the last two rows of \Cref{tab:cam_exp1}, using both Pl{\"u}cker ray and point clouds improves the results of very challenging camera trajectories.

\begin{table}
\centering
\caption{\textbf{Ablation of PCDController.} Different DiT layers are employed from top to bottom. PCDController adopts \textbf{\textcolor{blue!65}{20-layer}} as the final setting.\label{tab:ablation_study}}
\vspace{-0.15in}
\small
\begin{tabular}{cccccc}
\toprule 
\multicolumn{2}{c}{External DiT} & Overall & \multirow{2}{*}{ATE$\downarrow$} & \multirow{2}{*}{RPE$\downarrow$} & \multirow{2}{*}{RRE$\downarrow$}\tabularnewline
Layer & Param & Score &  &  & \tabularnewline
\midrule
10 & 0.48B & 87.75 & 0.136 & 0.0398 & 0.559\tabularnewline
\rowcolor{blue!20}
20 & 0.95B & \textbf{88.27} & \underline{0.102} & \underline{0.0313} & 0.246\tabularnewline
30 & 1.39B & \underline{88.08} & \textbf{0.092} & \textbf{0.0302} & \textbf{0.197}\tabularnewline
40 & 1.85B & 87.90 & 0.115 & 0.0339 & \underline{0.216}\tabularnewline
\bottomrule
\end{tabular}
\vspace{-0.1in}
\end{table}

\paragraph{Point Clouds of Humans.}
Human point clouds are always frozen in world space without any motion, which conflicts with the human motion conditions provided by SMPL-X.
Eliminating this ``redundant'' information is a straightforward idea to improve motion quality.
However, as verified in \Cref{tab:uni_exp1}, while Uni3C-T2V with human-masked point clouds achieves slightly better visual quality, this masking adversely affects camera precision, particularly when humans occupy a significant image area.
Therefore, retaining the point clouds of humans is essential for effective camera control. Given that Wan2.1~\cite{wang2025wan} is trained on videos featuring substantial motion, it can generate natural and smooth movements even when the foreground point clouds remain fixed.

\begin{figure}
\centering
\includegraphics[width=1.0\linewidth]{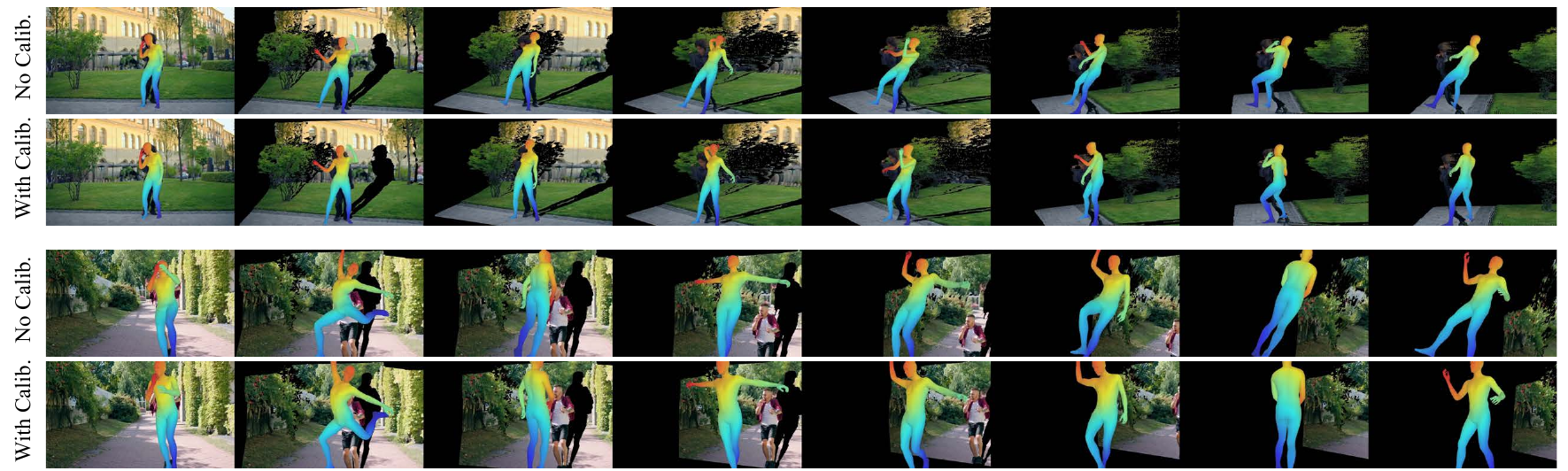}
\vspace{-0.15in}
   \caption{\textbf{Rendering results with and without gravity calibration by GeoCalib~\cite{veicht2024geocalib}.}
    \label{fig:gravity_calib}}
\vspace{-0.1in}
\end{figure}

\paragraph{Gravity Calibration.} 
As mentioned in \Cref{sec:3d_world}, gravity calibration is critical for aligning the global 3D world space. Results shown in \Cref{fig:gravity_calib} verify that the calibration can correct the SMPL-X characters aligned with skewed human point clouds and eliminate the error accumulation for humans' long-distance movements.

\begin{figure}
\centering
\includegraphics[width=1.0\linewidth]{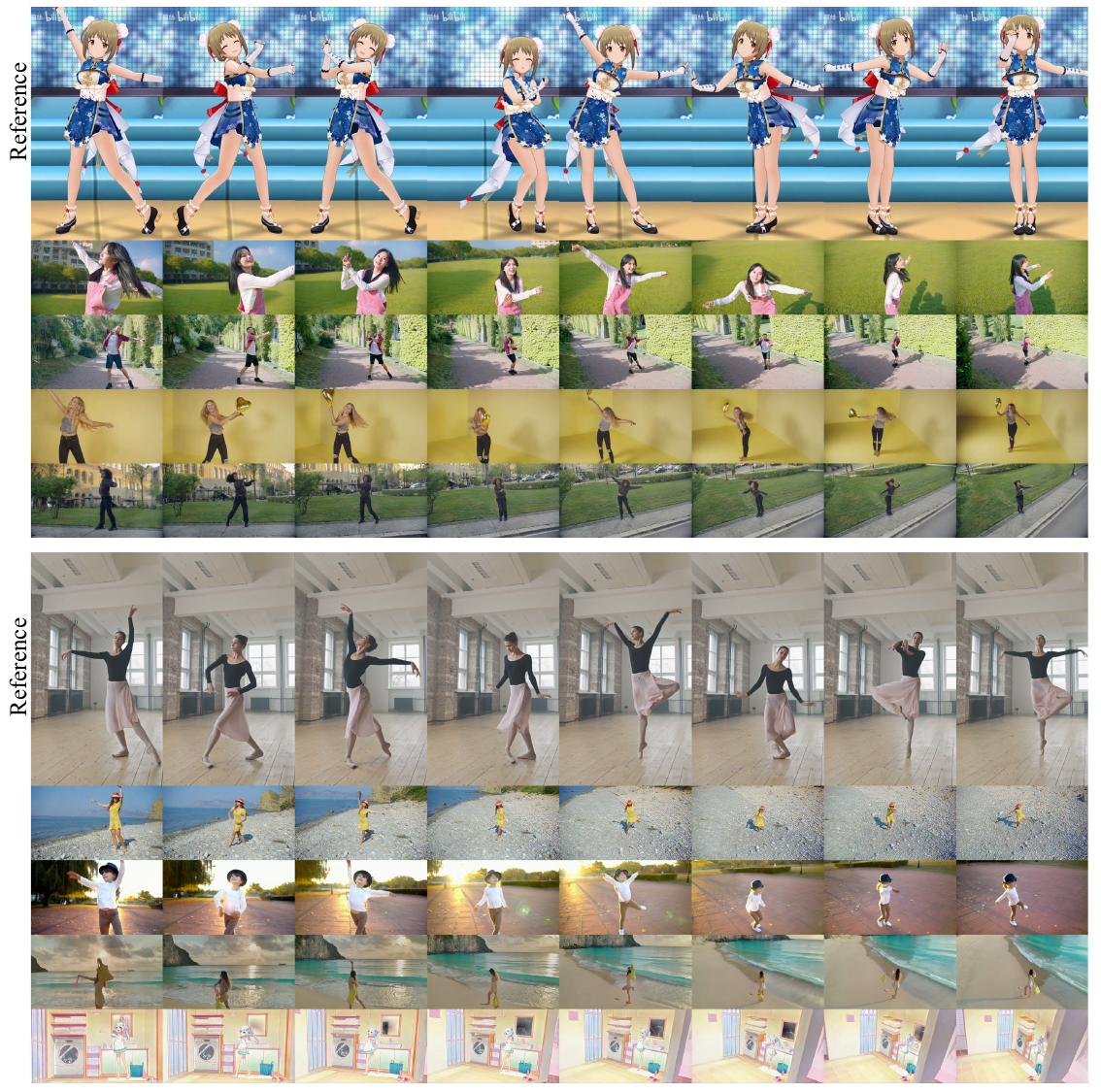}
\vspace{-0.15in}
   \caption{\textbf{Results of motion transfer.} The first row indicates the reference video, while others show our generated videos transferring motions from the reference sequence.
    \label{fig:motion_transfer}}
\vspace{-0.1in}
\end{figure}

\begin{figure}
\centering
\includegraphics[width=1.0\linewidth]{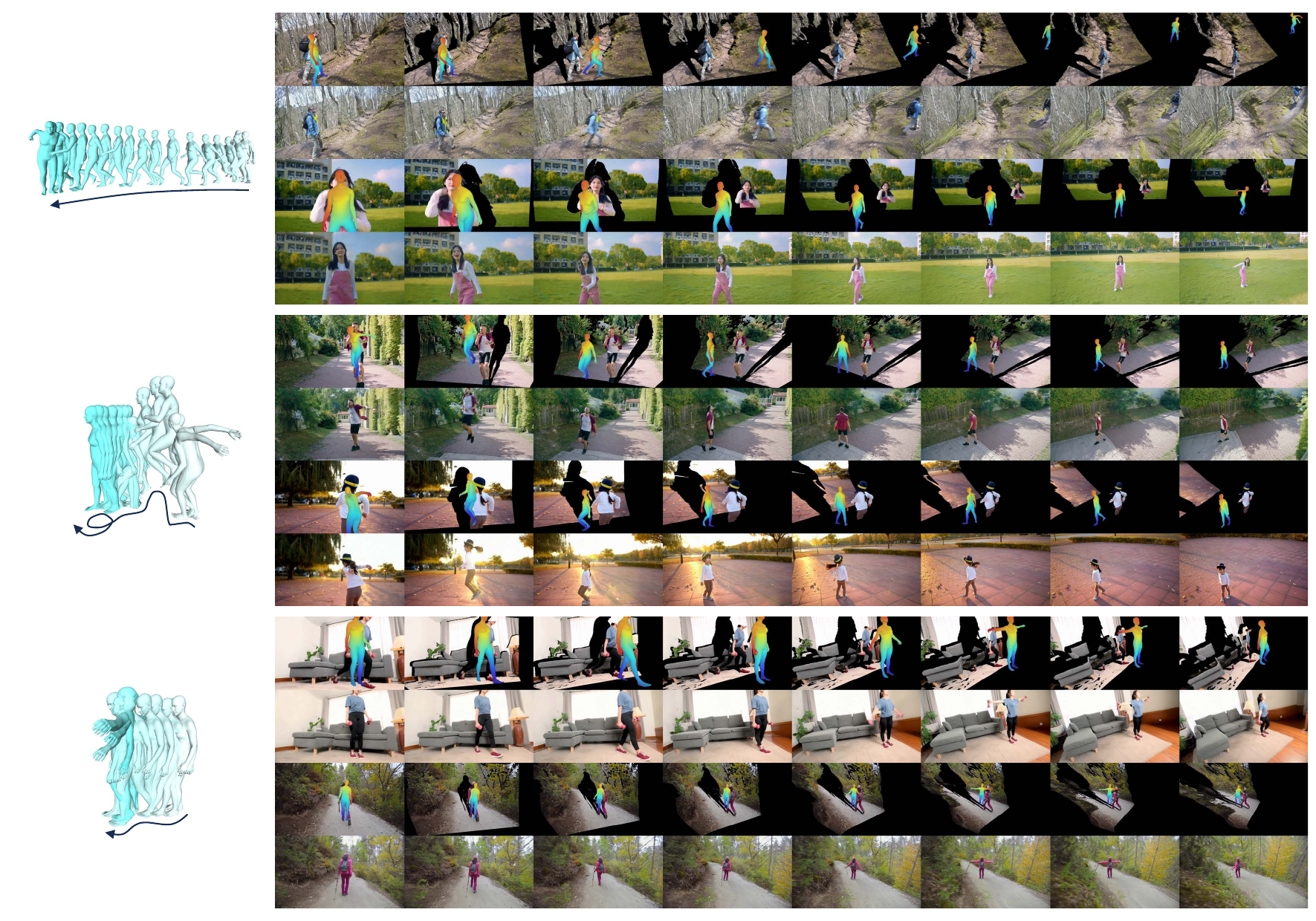}
 \vspace{-0.15in}
   \caption{\textbf{Results transferred from randomly integrated motion clips generated from text-to-motion~\cite{barquero2024seamless} trained on BABEL~\cite{punnakkal2021babel}.} The motion sequences are listed on the left, which are executed from light to dark colors. 
    \label{fig:motion_data_transfer}}
\vspace{-0.1in}
\end{figure}

\paragraph{Motion Transfer.}
We present motion transfer results achieved by the Uni3C framework in \Cref{fig:motion_transfer}.
Our model effectively controls both camera trajectories and human motions, even when reference motions are sourced from different videos or distinct domains, such as animation and real-world scenes.
Meanwhile, Uni3C can be further extended to generate vivid videos based on other conditions, like text-to-motion guidance or retrieved motions from motion databases. To prove this point, we randomly integrate several motion clips generated text-to-motion~\cite{barquero2024seamless} trained on BABEL~\cite{punnakkal2021babel} and use Uni3C to control both motion and camera, as illustrated in \Cref{fig:motion_data_transfer}. More results are shown in our supplementary.

\begin{figure}
\centering
\includegraphics[width=1.0\linewidth]{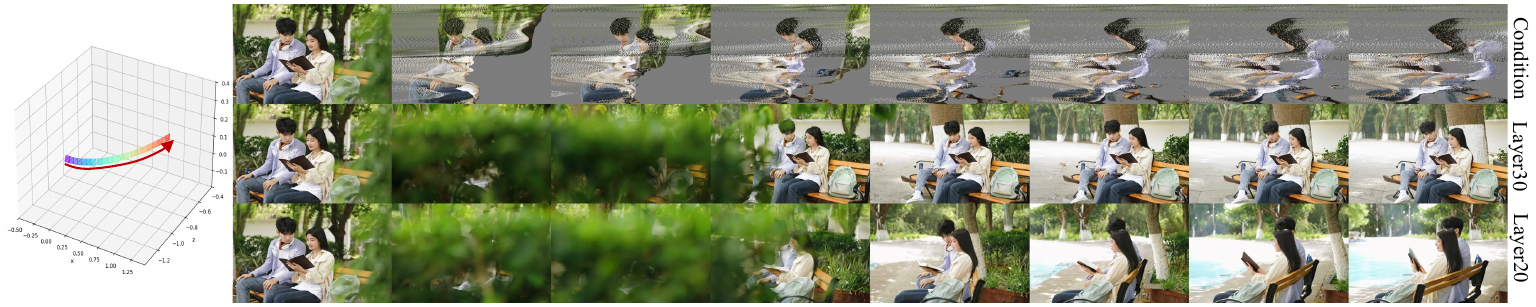}
\vspace{-0.15in}
   \caption{\textbf{Camera control results of PCDController with challenging point clouds.} The model with fewer external DiT layers enjoys superior conditional robustness compared to the one with more DiT layers. 
    \label{fig:ablation_vis}}
\vspace{-0.1in}
\end{figure}

\paragraph{Ablation of PCDController.}
We present some ablation results of PCDController with various external DiT layers in \Cref{tab:ablation_study}. Each setting injects conditional features into the main Wan-I2V backbone from top to bottom with different layers.
While the DiT branch with 30 layers shows slightly improved camera metrics, it ultimately compromises overall visual quality and generalization. We reveal that solely applying more external layers suffers from overfitting point cloud conditions as shown in \Cref{fig:ablation_vis}, resulting in suboptimal outcomes with distorted point clouds. Thus, we determine that employing 20 DiT layers achieves an effective balance between controllability and generalization within the PCDController.

\subsection{\revise{User Study}}

\revise{We conducted user studies for camera control (128 samples) and unified control (150 samples) on our respective benchmarks in \Cref{tab:user_study}. 
Formally, 20 unrelated volunteers are invited.
For the camera control, volunteers are requested to vote for the best result, considering video quality and camera precision, while the quality of human pose is additionally required for the unified control.
If two methods perform very similarly, volunteers are allowed to vote for both of them as top-2. We finally average the score for each metric.}

\begin{table}
\centering
\caption{\revise{\textbf{User studies on camera and unified (camera\&human) control.}\label{tab:user_study}}}
\vspace{-0.15in}
\footnotesize
\setlength{\tabcolsep}{2.5pt}
\begin{tabular}{l|llll}
\toprule
\textbf{Camera} & CamI2V & ViewCrafter & SEVA & PCDController\tabularnewline
\midrule
VideoQuality$\uparrow$ & 0.0391 & 0.0855 & \underline{0.2328} & \textbf{0.6457}\tabularnewline
CameraPrecision$\uparrow$ & 0.0359 & 0.1003 & \underline{0.2429} & \textbf{0.6281}\tabularnewline
\toprule
\textbf{Camera\&Human} & CamAnimate & RealisDance-DiT & PCDController & Uni3C\tabularnewline
\midrule
VideoQuality$\uparrow$ & 0.0170 & \underline{0.2466} & 0.0453 & \textbf{0.6920}\tabularnewline
CameraPrecision$\uparrow$ & 0.0676 & 0.2766 & \underline{0.5266} & \textbf{0.8976}\tabularnewline
HumanPose$\uparrow$ & 0.0716 & \underline{0.5863} & 0.1730 & \textbf{0.8983}\tabularnewline
\bottomrule
\end{tabular}
\vspace{-0.15in}
\end{table}

\section{Limitation and Future Work}

\begin{figure}[h]
\centering
\includegraphics[width=1.0\linewidth]{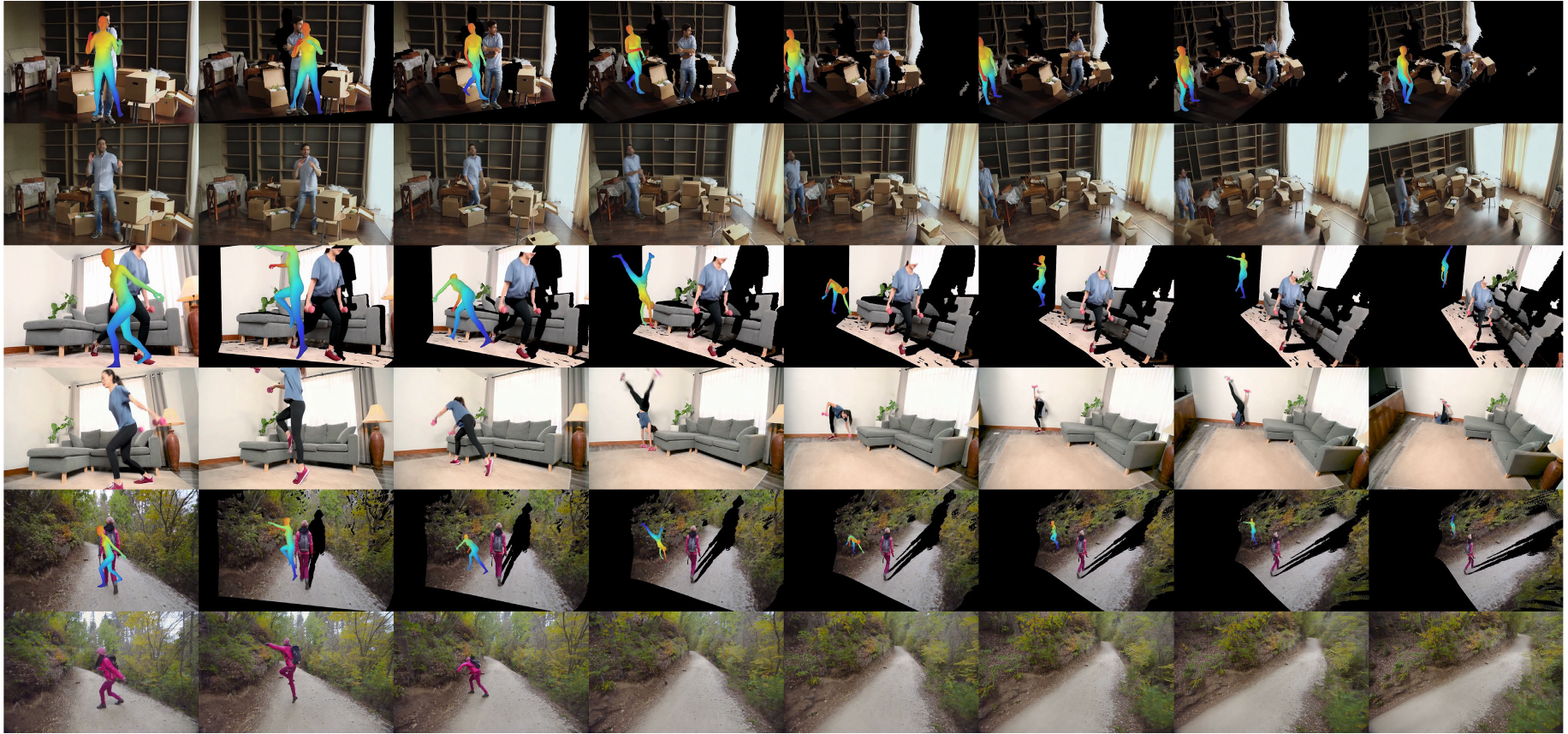}
\vspace{-0.2in}
  \caption{\textbf{Failed cases generated by Uni3C.} These results are primarily limited by the conflict between human motions and environments.
   \label{fig:failed_case}}
\vspace{-0.1in}
\end{figure}

\revise{Although Uni3C supports flexible and diverse unified control and motion transfer, it operates under the constraints of predefined camera trajectories and human characters (SMPL-X).
Consequently, Uni3C struggles to produce physically plausible outcomes when human motions conflict with environmental conditions, as shown in \Cref{fig:failed_case}.
For instance, if a human's movement trajectory is blocked by walls, barriers, or other objects, the generated results may exhibit artifacts such as distortion, clipping, or sudden disappearance.
This limitation could be mitigated by employing a more advanced human motion generation method that accounts for physical obstructions within the environment.
Besides, though our method can produce plausible results within inaccurate SMPL or world coordinates, it will result in undesired pose or motion direction.}

\section{Conclusion}

This paper introduced Uni3C, a framework that unifies 3D-enhanced camera and human motion controls for video generation. 
We first propose the PCDController, demonstrating that lightweight, trainable modules, and rich geometric priors from 3D point clouds can efficiently manage camera trajectories without compromising the inherent capacities of foundational VDMs. This not only enhances generalization but also facilitates versatile downstream applications without joint training.
Furthermore, by aligning multi-modal conditions, including both environmental point clouds and human characters in a global 3D world space, we established a coherent framework for jointly controlling camera movements and human animations. 
Our comprehensive experiments validated the efficacy of Uni3C across diverse datasets, showcasing superior performance in both quantitative and qualitative assessments compared to existing approaches.
The significance of our contributions lies not only in improving the state-of-the-art in controllable video generation but also in proposing a robust way to inject multi-modal conditions without the requirements of heavily annotated data. We believe that Uni3C paves the way for advanced controllable video generation.


\begin{acks}
This work was supported by the Science and Technology Commission of Shanghai Municipality (No. 24511103100).
\end{acks}

\clearpage 

\clearpage 

\clearpage

\bibliographystyle{ACM-Reference-Format}
\bibliography{sample-bibliography}


\begin{thebibliography}{79}


\ifx \showCODEN    \undefined \def \showCODEN     #1{\unskip}     \fi
\ifx \showISBNx    \undefined \def \showISBNx     #1{\unskip}     \fi
\ifx \showISBNxiii \undefined \def \showISBNxiii  #1{\unskip}     \fi
\ifx \showISSN     \undefined \def \showISSN      #1{\unskip}     \fi
\ifx \showLCCN     \undefined \def \showLCCN      #1{\unskip}     \fi
\ifx \shownote     \undefined \def \shownote      #1{#1}          \fi
\ifx \showarticletitle \undefined \def \showarticletitle #1{#1}   \fi
\ifx \showURL      \undefined \def \showURL       {\relax}        \fi
\providecommand\bibfield[2]{#2}
\providecommand\bibinfo[2]{#2}
\providecommand\natexlab[1]{#1}
\providecommand\showeprint[2][]{arXiv:#2}

\bibitem[Arnold et~al\mbox{.}(2022)]%
        {arnold2022map}
\bibfield{author}{\bibinfo{person}{Eduardo Arnold}, \bibinfo{person}{Jamie
  Wynn}, \bibinfo{person}{Sara Vicente}, \bibinfo{person}{Guillermo
  Garcia-Hernando}, \bibinfo{person}{Aron Monszpart}, \bibinfo{person}{Victor
  Prisacariu}, \bibinfo{person}{Daniyar Turmukhambetov}, {and}
  \bibinfo{person}{Eric Brachmann}.} \bibinfo{year}{2022}\natexlab{}.
\newblock \showarticletitle{Map-free visual relocalization: Metric pose
  relative to a single image}. In \bibinfo{booktitle}{\emph{European Conference
  on Computer Vision}}. Springer, \bibinfo{pages}{690--708}.
\newblock


\bibitem[Bahmani et~al\mbox{.}(2025a)]%
        {bahmani2024ac3d}
\bibfield{author}{\bibinfo{person}{Sherwin Bahmani}, \bibinfo{person}{Ivan
  Skorokhodov}, \bibinfo{person}{Guocheng Qian}, \bibinfo{person}{Aliaksandr
  Siarohin}, \bibinfo{person}{Willi Menapace}, \bibinfo{person}{Andrea
  Tagliasacchi}, \bibinfo{person}{David~B Lindell}, {and}
  \bibinfo{person}{Sergey Tulyakov}.} \bibinfo{year}{2025}\natexlab{a}.
\newblock \showarticletitle{AC3D: Analyzing and Improving 3D Camera Control in
  Video Diffusion Transformers}. In \bibinfo{booktitle}{\emph{Proceedings of
  the IEEE/CVF Conference on Computer Vision and Pattern Recognition}}.
\newblock


\bibitem[Bahmani et~al\mbox{.}(2025b)]%
        {bahmani2024vd3d}
\bibfield{author}{\bibinfo{person}{Sherwin Bahmani}, \bibinfo{person}{Ivan
  Skorokhodov}, \bibinfo{person}{Aliaksandr Siarohin}, \bibinfo{person}{Willi
  Menapace}, \bibinfo{person}{Guocheng Qian}, \bibinfo{person}{Michael
  Vasilkovsky}, \bibinfo{person}{Hsin-Ying Lee}, \bibinfo{person}{Chaoyang
  Wang}, \bibinfo{person}{Jiaxu Zou}, \bibinfo{person}{Andrea Tagliasacchi},
  {et~al\mbox{.}}} \bibinfo{year}{2025}\natexlab{b}.
\newblock \showarticletitle{Vd3d: Taming large video diffusion transformers for
  3d camera control}. In \bibinfo{booktitle}{\emph{International Conference on
  Learning Representations}}.
\newblock


\bibitem[Barquero et~al\mbox{.}(2024)]%
        {barquero2024seamless}
\bibfield{author}{\bibinfo{person}{German Barquero}, \bibinfo{person}{Sergio
  Escalera}, {and} \bibinfo{person}{Cristina Palmero}.}
  \bibinfo{year}{2024}\natexlab{}.
\newblock \showarticletitle{Seamless Human Motion Composition with Blended
  Positional Encodings}. In \bibinfo{booktitle}{\emph{Proceedings of the
  IEEE/CVF Conference on Computer Vision and Pattern Recognition}}.
\newblock


\bibitem[Barron et~al\mbox{.}(2022)]%
        {barron2022mip}
\bibfield{author}{\bibinfo{person}{Jonathan~T Barron}, \bibinfo{person}{Ben
  Mildenhall}, \bibinfo{person}{Dor Verbin}, \bibinfo{person}{Pratul~P
  Srinivasan}, {and} \bibinfo{person}{Peter Hedman}.}
  \bibinfo{year}{2022}\natexlab{}.
\newblock \showarticletitle{Mip-nerf 360: Unbounded anti-aliased neural
  radiance fields}. In \bibinfo{booktitle}{\emph{Proceedings of the IEEE/CVF
  conference on computer vision and pattern recognition}}.
  \bibinfo{pages}{5470--5479}.
\newblock


\bibitem[Blattmann et~al\mbox{.}(2023)]%
        {Blattmann2023svd}
\bibfield{author}{\bibinfo{person}{Andreas Blattmann}, \bibinfo{person}{Tim
  Dockhorn}, \bibinfo{person}{Sumith Kulal}, \bibinfo{person}{Daniel
  Mendelevitch}, \bibinfo{person}{Maciej Kilian}, \bibinfo{person}{Dominik
  Lorenz}, \bibinfo{person}{Yam Levi}, \bibinfo{person}{Zion English},
  \bibinfo{person}{Vikram Voleti}, \bibinfo{person}{Adam Letts},
  {et~al\mbox{.}}} \bibinfo{year}{2023}\natexlab{}.
\newblock \showarticletitle{Stable video diffusion: Scaling latent video
  diffusion models to large datasets}.
\newblock \bibinfo{journal}{\emph{arXiv preprint arXiv:2311.15127}}
  (\bibinfo{year}{2023}).
\newblock


\bibitem[Bochkovskii et~al\mbox{.}(2024)]%
        {Bochkovskii2024depthpro}
\bibfield{author}{\bibinfo{person}{Aleksei Bochkovskii},
  \bibinfo{person}{Ama\"{e}l Delaunoy}, \bibinfo{person}{Hugo Germain},
  \bibinfo{person}{Marcel Santos}, \bibinfo{person}{Yichao Zhou},
  \bibinfo{person}{Stephan~R. Richter}, {and} \bibinfo{person}{Vladlen
  Koltun}.} \bibinfo{year}{2024}\natexlab{}.
\newblock \showarticletitle{Depth Pro: Sharp Monocular Metric Depth in Less
  Than a Second}.
\newblock \bibinfo{journal}{\emph{arXiv}} (\bibinfo{year}{2024}).
\newblock
\urldef\tempurl%
\url{https://arxiv.org/abs/2410.02073}
\showURL{%
\tempurl}


\bibitem[Brooks et~al\mbox{.}(2024)]%
        {Tim2024sora}
\bibfield{author}{\bibinfo{person}{Tim Brooks}, \bibinfo{person}{Bill Peebles},
  \bibinfo{person}{Connor Holmes}, \bibinfo{person}{Will DePue},
  \bibinfo{person}{Yufei Guo}, \bibinfo{person}{Li Jing},
  \bibinfo{person}{David Schnurr}, \bibinfo{person}{Joe Taylor},
  \bibinfo{person}{Troy Luhman}, \bibinfo{person}{Eric Luhman},
  \bibinfo{person}{Clarence Ng}, \bibinfo{person}{Ricky Wang}, {and}
  \bibinfo{person}{Aditya Ramesh}.} \bibinfo{year}{2024}\natexlab{}.
\newblock \showarticletitle{Video generation models as world simulators}.
\newblock  (\bibinfo{year}{2024}).
\newblock
\urldef\tempurl%
\url{https://openai.com/research/video-generation-models-as-world-simulators}
\showURL{%
\tempurl}


\bibitem[Cao et~al\mbox{.}(2024)]%
        {cao2024mvsformer++}
\bibfield{author}{\bibinfo{person}{Chenjie Cao}, \bibinfo{person}{Xinlin Ren},
  {and} \bibinfo{person}{Yanwei Fu}.} \bibinfo{year}{2024}\natexlab{}.
\newblock \showarticletitle{MVSFormer++: Revealing the Devil in Transformer's
  Details for Multi-View Stereo}. In \bibinfo{booktitle}{\emph{International
  Conference on Learning Representations}}.
\newblock


\bibitem[Cao et~al\mbox{.}(2025)]%
        {cao2024mvgenmaster}
\bibfield{author}{\bibinfo{person}{Chenjie Cao}, \bibinfo{person}{Chaohui Yu},
  \bibinfo{person}{Shang Liu}, \bibinfo{person}{Fan Wang},
  \bibinfo{person}{Xiangyang Xue}, {and} \bibinfo{person}{Yanwei Fu}.}
  \bibinfo{year}{2025}\natexlab{}.
\newblock \showarticletitle{MVGenMaster: Scaling Multi-View Generation from Any
  Image via 3D Priors Enhanced Diffusion Model}. In
  \bibinfo{booktitle}{\emph{Proceedings of the IEEE/CVF Conference on Computer
  Vision and Pattern Recognition}}.
\newblock


\bibitem[Chen et~al\mbox{.}(2025)]%
        {chen2025perception}
\bibfield{author}{\bibinfo{person}{Yingjie Chen}, \bibinfo{person}{Yifang Men},
  \bibinfo{person}{Yuan Yao}, \bibinfo{person}{Miaomiao Cui}, {and}
  \bibinfo{person}{Liefeng Bo}.} \bibinfo{year}{2025}\natexlab{}.
\newblock \showarticletitle{Perception-as-Control: Fine-grained Controllable
  Image Animation with 3D-aware Motion Representation}.
\newblock \bibinfo{journal}{\emph{arXiv preprint arXiv:2501.05020}}
  (\bibinfo{year}{2025}).
\newblock


\bibitem[Chung et~al\mbox{.}(2023)]%
        {chung2023unimax}
\bibfield{author}{\bibinfo{person}{Hyung~Won Chung}, \bibinfo{person}{Noah
  Constant}, \bibinfo{person}{Xavier Garcia}, \bibinfo{person}{Adam Roberts},
  \bibinfo{person}{Yi Tay}, \bibinfo{person}{Sharan Narang}, {and}
  \bibinfo{person}{Orhan Firat}.} \bibinfo{year}{2023}\natexlab{}.
\newblock \showarticletitle{Unimax: Fairer and more effective language sampling
  for large-scale multilingual pretraining}.
\newblock \bibinfo{journal}{\emph{arXiv preprint arXiv:2304.09151}}
  (\bibinfo{year}{2023}).
\newblock


\bibitem[Feng et~al\mbox{.}(2025)]%
        {feng2024i2vcontrol}
\bibfield{author}{\bibinfo{person}{Wanquan Feng}, \bibinfo{person}{Jiawei Liu},
  \bibinfo{person}{Pengqi Tu}, \bibinfo{person}{Tianhao Qi},
  \bibinfo{person}{Mingzhen Sun}, \bibinfo{person}{Tianxiang Ma},
  \bibinfo{person}{Songtao Zhao}, \bibinfo{person}{Siyu Zhou}, {and}
  \bibinfo{person}{Qian He}.} \bibinfo{year}{2025}\natexlab{}.
\newblock \showarticletitle{I2VControl-Camera: Precise Video Camera Control
  with Adjustable Motion Strength}. In \bibinfo{booktitle}{\emph{International
  Conference on Learning Representations}}.
\newblock


\bibitem[Geng et~al\mbox{.}(2025)]%
        {geng2024motionprompting}
\bibfield{author}{\bibinfo{person}{Daniel Geng}, \bibinfo{person}{Charles
  Herrmann}, \bibinfo{person}{Junhwa Hur}, \bibinfo{person}{Forrester Cole},
  \bibinfo{person}{Serena Zhang}, \bibinfo{person}{Tobias Pfaff},
  \bibinfo{person}{Tatiana Lopez-Guevara}, \bibinfo{person}{Carl Doersch},
  \bibinfo{person}{Yusuf Aytar}, \bibinfo{person}{Michael Rubinstein},
  \bibinfo{person}{Chen Sun}, \bibinfo{person}{Oliver Wang},
  \bibinfo{person}{Andrew Owens}, {and} \bibinfo{person}{Deqing Sun}.}
  \bibinfo{year}{2025}\natexlab{}.
\newblock \showarticletitle{Motion Prompting: Controlling Video Generation with
  Motion Trajectories}. In \bibinfo{booktitle}{\emph{Proceedings of the
  IEEE/CVF conference on computer vision and pattern recognition}}.
\newblock


\bibitem[Gu et~al\mbox{.}(2025)]%
        {gu2025diffusion}
\bibfield{author}{\bibinfo{person}{Zekai Gu}, \bibinfo{person}{Rui Yan},
  \bibinfo{person}{Jiahao Lu}, \bibinfo{person}{Peng Li},
  \bibinfo{person}{Zhiyang Dou}, \bibinfo{person}{Chenyang Si},
  \bibinfo{person}{Zhen Dong}, \bibinfo{person}{Qifeng Liu},
  \bibinfo{person}{Cheng Lin}, \bibinfo{person}{Ziwei Liu}, {et~al\mbox{.}}}
  \bibinfo{year}{2025}\natexlab{}.
\newblock \showarticletitle{Diffusion as Shader: 3D-aware Video Diffusion for
  Versatile Video Generation Control}.
\newblock \bibinfo{journal}{\emph{arXiv preprint arXiv:2501.03847}}
  (\bibinfo{year}{2025}).
\newblock


\bibitem[Guo et~al\mbox{.}(2024)]%
        {guo2024momask}
\bibfield{author}{\bibinfo{person}{Chuan Guo}, \bibinfo{person}{Yuxuan Mu},
  \bibinfo{person}{Muhammad~Gohar Javed}, \bibinfo{person}{Sen Wang}, {and}
  \bibinfo{person}{Li Cheng}.} \bibinfo{year}{2024}\natexlab{}.
\newblock \showarticletitle{Momask: Generative masked modeling of 3d human
  motions}. In \bibinfo{booktitle}{\emph{Proceedings of the IEEE/CVF Conference
  on Computer Vision and Pattern Recognition}}. \bibinfo{pages}{1900--1910}.
\newblock


\bibitem[Guo et~al\mbox{.}(2022)]%
        {guo2022generating}
\bibfield{author}{\bibinfo{person}{Chuan Guo}, \bibinfo{person}{Shihao Zou},
  \bibinfo{person}{Xinxin Zuo}, \bibinfo{person}{Sen Wang},
  \bibinfo{person}{Wei Ji}, \bibinfo{person}{Xingyu Li}, {and}
  \bibinfo{person}{Li Cheng}.} \bibinfo{year}{2022}\natexlab{}.
\newblock \showarticletitle{Generating diverse and natural 3d human motions
  from text}. In \bibinfo{booktitle}{\emph{Proceedings of the IEEE/CVF
  conference on computer vision and pattern recognition}}.
  \bibinfo{pages}{5152--5161}.
\newblock


\bibitem[He et~al\mbox{.}(2025a)]%
        {He2024Cameractrl}
\bibfield{author}{\bibinfo{person}{Hao He}, \bibinfo{person}{Yinghao Xu},
  \bibinfo{person}{Yuwei Guo}, \bibinfo{person}{Gordon Wetzstein},
  \bibinfo{person}{Bo Dai}, \bibinfo{person}{Hongsheng Li}, {and}
  \bibinfo{person}{Ceyuan Yang}.} \bibinfo{year}{2025}\natexlab{a}.
\newblock \showarticletitle{Cameractrl: Enabling camera control for
  text-to-video generation}. In \bibinfo{booktitle}{\emph{International
  Conference on Learning Representations}}.
\newblock


\bibitem[He et~al\mbox{.}(2025b)]%
        {he2025cameractrl}
\bibfield{author}{\bibinfo{person}{Hao He}, \bibinfo{person}{Ceyuan Yang},
  \bibinfo{person}{Shanchuan Lin}, \bibinfo{person}{Yinghao Xu},
  \bibinfo{person}{Meng Wei}, \bibinfo{person}{Liangke Gui},
  \bibinfo{person}{Qi Zhao}, \bibinfo{person}{Gordon Wetzstein},
  \bibinfo{person}{Lu Jiang}, {and} \bibinfo{person}{Hongsheng Li}.}
  \bibinfo{year}{2025}\natexlab{b}.
\newblock \showarticletitle{CameraCtrl II: Dynamic Scene Exploration via
  Camera-controlled Video Diffusion Models}.
\newblock \bibinfo{journal}{\emph{arXiv preprint arXiv:2503.10592}}
  (\bibinfo{year}{2025}).
\newblock


\bibitem[Hou et~al\mbox{.}(2025)]%
        {hou2024training}
\bibfield{author}{\bibinfo{person}{Chen Hou}, \bibinfo{person}{Guoqiang Wei},
  \bibinfo{person}{Yan Zeng}, {and} \bibinfo{person}{Zhibo Chen}.}
  \bibinfo{year}{2025}\natexlab{}.
\newblock \showarticletitle{Training-free camera control for video generation}.
  In \bibinfo{booktitle}{\emph{International Conference on Learning
  Representations}}.
\newblock


\bibitem[Hu et~al\mbox{.}(2022)]%
        {hu2021lora}
\bibfield{author}{\bibinfo{person}{Edward~J Hu}, \bibinfo{person}{Yelong Shen},
  \bibinfo{person}{Phillip Wallis}, \bibinfo{person}{Zeyuan Allen-Zhu},
  \bibinfo{person}{Yuanzhi Li}, \bibinfo{person}{Shean Wang},
  \bibinfo{person}{Lu Wang}, {and} \bibinfo{person}{Weizhu Chen}.}
  \bibinfo{year}{2022}\natexlab{}.
\newblock \showarticletitle{Lora: Low-rank adaptation of large language
  models}. In \bibinfo{booktitle}{\emph{International Conference on Learning
  Representations}}.
\newblock


\bibitem[Hu(2024)]%
        {hu2024animate}
\bibfield{author}{\bibinfo{person}{Li Hu}.} \bibinfo{year}{2024}\natexlab{}.
\newblock \showarticletitle{Animate anyone: Consistent and controllable
  image-to-video synthesis for character animation}. In
  \bibinfo{booktitle}{\emph{Proceedings of the IEEE/CVF Conference on Computer
  Vision and Pattern Recognition}}. \bibinfo{pages}{8153--8163}.
\newblock


\bibitem[Hu et~al\mbox{.}(2025)]%
        {animateanyone2}
\bibfield{author}{\bibinfo{person}{Li Hu}, \bibinfo{person}{Guangyuan Wang},
  \bibinfo{person}{Zhen Shen}, \bibinfo{person}{Xin Gao},
  \bibinfo{person}{Dechao Meng}, \bibinfo{person}{Lian Zhuo},
  \bibinfo{person}{Peng Zhang}, \bibinfo{person}{Bang Zhang}, {and}
  \bibinfo{person}{Liefeng Bo}.} \bibinfo{year}{2025}\natexlab{}.
\newblock \showarticletitle{Animate Anyone 2: High-Fidelity Character Image
  Animation with Environment Affordance}.
\newblock \bibinfo{journal}{\emph{arXiv preprint arXiv:2502.06145}}
  (\bibinfo{year}{2025}).
\newblock


\bibitem[Huang et~al\mbox{.}(2024)]%
        {huang2024vbench++}
\bibfield{author}{\bibinfo{person}{Ziqi Huang}, \bibinfo{person}{Fan Zhang},
  \bibinfo{person}{Xiaojie Xu}, \bibinfo{person}{Yinan He},
  \bibinfo{person}{Jiashuo Yu}, \bibinfo{person}{Ziyue Dong},
  \bibinfo{person}{Qianli Ma}, \bibinfo{person}{Nattapol Chanpaisit},
  \bibinfo{person}{Chenyang Si}, \bibinfo{person}{Yuming Jiang},
  {et~al\mbox{.}}} \bibinfo{year}{2024}\natexlab{}.
\newblock \showarticletitle{Vbench++: Comprehensive and versatile benchmark
  suite for video generative models}.
\newblock \bibinfo{journal}{\emph{arXiv preprint arXiv:2411.13503}}
  (\bibinfo{year}{2024}).
\newblock


\bibitem[Jiang et~al\mbox{.}(2023)]%
        {jiang2023motiongpt}
\bibfield{author}{\bibinfo{person}{Biao Jiang}, \bibinfo{person}{Xin Chen},
  \bibinfo{person}{Wen Liu}, \bibinfo{person}{Jingyi Yu}, \bibinfo{person}{Gang
  Yu}, {and} \bibinfo{person}{Tao Chen}.} \bibinfo{year}{2023}\natexlab{}.
\newblock \showarticletitle{Motiongpt: Human motion as a foreign language}.
\newblock \bibinfo{journal}{\emph{Advances in Neural Information Processing
  Systems}}  \bibinfo{volume}{36} (\bibinfo{year}{2023}),
  \bibinfo{pages}{20067--20079}.
\newblock


\bibitem[Jiang et~al\mbox{.}(2024)]%
        {jiang2024cinematic}
\bibfield{author}{\bibinfo{person}{Xuekun Jiang}, \bibinfo{person}{Anyi Rao},
  \bibinfo{person}{Jingbo Wang}, \bibinfo{person}{Dahua Lin}, {and}
  \bibinfo{person}{Bo Dai}.} \bibinfo{year}{2024}\natexlab{}.
\newblock \showarticletitle{Cinematic behavior transfer via nerf-based
  differentiable filming}. In \bibinfo{booktitle}{\emph{Proceedings of the
  IEEE/CVF Conference on Computer Vision and Pattern Recognition}}.
  \bibinfo{pages}{6723--6732}.
\newblock


\bibitem[Knapitsch et~al\mbox{.}(2017)]%
        {knapitsch2017tanks}
\bibfield{author}{\bibinfo{person}{Arno Knapitsch}, \bibinfo{person}{Jaesik
  Park}, \bibinfo{person}{Qian-Yi Zhou}, {and} \bibinfo{person}{Vladlen
  Koltun}.} \bibinfo{year}{2017}\natexlab{}.
\newblock \showarticletitle{Tanks and temples: Benchmarking large-scale scene
  reconstruction}.
\newblock \bibinfo{journal}{\emph{ACM Transactions on Graphics (ToG)}}
  \bibinfo{volume}{36}, \bibinfo{number}{4} (\bibinfo{year}{2017}),
  \bibinfo{pages}{1--13}.
\newblock


\bibitem[Kocabas et~al\mbox{.}(2024)]%
        {kocabas2024pace}
\bibfield{author}{\bibinfo{person}{Muhammed Kocabas}, \bibinfo{person}{Ye
  Yuan}, \bibinfo{person}{Pavlo Molchanov}, \bibinfo{person}{Yunrong Guo},
  \bibinfo{person}{Michael~J Black}, \bibinfo{person}{Otmar Hilliges},
  \bibinfo{person}{Jan Kautz}, {and} \bibinfo{person}{Umar Iqbal}.}
  \bibinfo{year}{2024}\natexlab{}.
\newblock \showarticletitle{PACE: Human and Camera Motion Estimation from
  in-the-wild Videos}. In \bibinfo{booktitle}{\emph{2024 International
  Conference on 3D Vision (3DV)}}. IEEE, \bibinfo{pages}{397--408}.
\newblock


\bibitem[Kong et~al\mbox{.}(2024)]%
        {kong2024hunyuanvideo}
\bibfield{author}{\bibinfo{person}{Weijie Kong}, \bibinfo{person}{Qi Tian},
  \bibinfo{person}{Zijian Zhang}, \bibinfo{person}{Rox Min},
  \bibinfo{person}{Zuozhuo Dai}, \bibinfo{person}{Jin Zhou},
  \bibinfo{person}{Jiangfeng Xiong}, \bibinfo{person}{Xin Li},
  \bibinfo{person}{Bo Wu}, \bibinfo{person}{Jianwei Zhang}, {et~al\mbox{.}}}
  \bibinfo{year}{2024}\natexlab{}.
\newblock \showarticletitle{Hunyuanvideo: A systematic framework for large
  video generative models}.
\newblock \bibinfo{journal}{\emph{arXiv preprint arXiv:2412.03603}}
  (\bibinfo{year}{2024}).
\newblock


\bibitem[Kuaishou(2024)]%
        {kling}
\bibfield{author}{\bibinfo{person}{Kuaishou}.} \bibinfo{year}{2024}\natexlab{}.
\newblock \showarticletitle{Kling}.
\newblock \bibinfo{journal}{\emph{https://klingai.kuaishou.com}}
  (\bibinfo{year}{2024}).
\newblock


\bibitem[Li et~al\mbox{.}(2025b)]%
        {li2025realcam}
\bibfield{author}{\bibinfo{person}{Teng Li}, \bibinfo{person}{Guangcong Zheng},
  \bibinfo{person}{Rui Jiang}, \bibinfo{person}{Tao Wu}, \bibinfo{person}{Yehao
  Lu}, \bibinfo{person}{Yining Lin}, \bibinfo{person}{Xi Li}, {et~al\mbox{.}}}
  \bibinfo{year}{2025}\natexlab{b}.
\newblock \showarticletitle{RealCam-I2V: Real-World Image-to-Video Generation
  with Interactive Complex Camera Control}.
\newblock \bibinfo{journal}{\emph{arXiv preprint arXiv:2502.10059}}
  (\bibinfo{year}{2025}).
\newblock


\bibitem[Li et~al\mbox{.}(2025a)]%
        {li2024megasam}
\bibfield{author}{\bibinfo{person}{Zhengqi Li}, \bibinfo{person}{Richard
  Tucker}, \bibinfo{person}{Forrester Cole}, \bibinfo{person}{Qianqian Wang},
  \bibinfo{person}{Linyi Jin}, \bibinfo{person}{Vickie Ye},
  \bibinfo{person}{Angjoo Kanazawa}, \bibinfo{person}{Aleksander Holynski},
  {and} \bibinfo{person}{Noah Snavely}.} \bibinfo{year}{2025}\natexlab{a}.
\newblock \showarticletitle{Megasam: Accurate, fast, and robust structure and
  motion from casual dynamic videos}. In \bibinfo{booktitle}{\emph{Proceedings
  of the IEEE/CVF Conference on Computer Vision and Pattern Recognition}}.
\newblock


\bibitem[Liang et~al\mbox{.}(2024)]%
        {liang2024wonderland}
\bibfield{author}{\bibinfo{person}{Hanwen Liang}, \bibinfo{person}{Junli Cao},
  \bibinfo{person}{Vidit Goel}, \bibinfo{person}{Guocheng Qian},
  \bibinfo{person}{Sergei Korolev}, \bibinfo{person}{Demetri Terzopoulos},
  \bibinfo{person}{Konstantinos~N Plataniotis}, \bibinfo{person}{Sergey
  Tulyakov}, {and} \bibinfo{person}{Jian Ren}.}
  \bibinfo{year}{2024}\natexlab{}.
\newblock \showarticletitle{Wonderland: Navigating 3D Scenes from a Single
  Image}.
\newblock \bibinfo{journal}{\emph{arXiv preprint arXiv:2412.12091}}
  (\bibinfo{year}{2024}).
\newblock


\bibitem[Lin et~al\mbox{.}(2023)]%
        {lin2023motion}
\bibfield{author}{\bibinfo{person}{Jing Lin}, \bibinfo{person}{Ailing Zeng},
  \bibinfo{person}{Shunlin Lu}, \bibinfo{person}{Yuanhao Cai},
  \bibinfo{person}{Ruimao Zhang}, \bibinfo{person}{Haoqian Wang}, {and}
  \bibinfo{person}{Lei Zhang}.} \bibinfo{year}{2023}\natexlab{}.
\newblock \showarticletitle{Motion-x: A large-scale 3d expressive whole-body
  human motion dataset}.
\newblock \bibinfo{journal}{\emph{Advances in Neural Information Processing
  Systems}}  \bibinfo{volume}{36} (\bibinfo{year}{2023}),
  \bibinfo{pages}{25268--25280}.
\newblock


\bibitem[Ling et~al\mbox{.}(2024)]%
        {ling2024dl3dv}
\bibfield{author}{\bibinfo{person}{Lu Ling}, \bibinfo{person}{Yichen Sheng},
  \bibinfo{person}{Zhi Tu}, \bibinfo{person}{Wentian Zhao},
  \bibinfo{person}{Cheng Xin}, \bibinfo{person}{Kun Wan},
  \bibinfo{person}{Lantao Yu}, \bibinfo{person}{Qianyu Guo},
  \bibinfo{person}{Zixun Yu}, \bibinfo{person}{Yawen Lu}, {et~al\mbox{.}}}
  \bibinfo{year}{2024}\natexlab{}.
\newblock \showarticletitle{Dl3dv-10k: A large-scale scene dataset for deep
  learning-based 3d vision}. In \bibinfo{booktitle}{\emph{Proceedings of the
  IEEE/CVF Conference on Computer Vision and Pattern Recognition}}.
  \bibinfo{pages}{22160--22169}.
\newblock


\bibitem[Lipman et~al\mbox{.}(2022)]%
        {lipman2022flow}
\bibfield{author}{\bibinfo{person}{Yaron Lipman}, \bibinfo{person}{Ricky~TQ
  Chen}, \bibinfo{person}{Heli Ben-Hamu}, \bibinfo{person}{Maximilian Nickel},
  {and} \bibinfo{person}{Matt Le}.} \bibinfo{year}{2022}\natexlab{}.
\newblock \showarticletitle{Flow matching for generative modeling}.
\newblock \bibinfo{journal}{\emph{arXiv preprint arXiv:2210.02747}}
  (\bibinfo{year}{2022}).
\newblock


\bibitem[Liu et~al\mbox{.}(2021)]%
        {liu2021infinite}
\bibfield{author}{\bibinfo{person}{Andrew Liu}, \bibinfo{person}{Richard
  Tucker}, \bibinfo{person}{Varun Jampani}, \bibinfo{person}{Ameesh Makadia},
  \bibinfo{person}{Noah Snavely}, {and} \bibinfo{person}{Angjoo Kanazawa}.}
  \bibinfo{year}{2021}\natexlab{}.
\newblock \showarticletitle{Infinite nature: Perpetual view generation of
  natural scenes from a single image}. In \bibinfo{booktitle}{\emph{Proceedings
  of the IEEE/CVF International Conference on Computer Vision}}.
  \bibinfo{pages}{14458--14467}.
\newblock


\bibitem[Liu et~al\mbox{.}(2025)]%
        {liu2025uncommon}
\bibfield{author}{\bibinfo{person}{Xingchen Liu}, \bibinfo{person}{Piyush
  Tayal}, \bibinfo{person}{Jianyuan Wang}, \bibinfo{person}{Jesus Zarzar},
  \bibinfo{person}{Tom Monnier}, \bibinfo{person}{Konstantinos Tertikas},
  \bibinfo{person}{Jiali Duan}, \bibinfo{person}{Antoine Toisoul},
  \bibinfo{person}{Jason~Y Zhang}, \bibinfo{person}{Natalia Neverova},
  {et~al\mbox{.}}} \bibinfo{year}{2025}\natexlab{}.
\newblock \showarticletitle{UnCommon Objects in 3D}.
\newblock \bibinfo{journal}{\emph{arXiv preprint arXiv:2501.07574}}
  (\bibinfo{year}{2025}).
\newblock


\bibitem[Pavlakos et~al\mbox{.}(2019)]%
        {pavlakos2019expressive}
\bibfield{author}{\bibinfo{person}{Georgios Pavlakos},
  \bibinfo{person}{Vasileios Choutas}, \bibinfo{person}{Nima Ghorbani},
  \bibinfo{person}{Timo Bolkart}, \bibinfo{person}{Ahmed~AA Osman},
  \bibinfo{person}{Dimitrios Tzionas}, {and} \bibinfo{person}{Michael~J
  Black}.} \bibinfo{year}{2019}\natexlab{}.
\newblock \showarticletitle{Expressive body capture: 3d hands, face, and body
  from a single image}. In \bibinfo{booktitle}{\emph{Proceedings of the
  IEEE/CVF Conference on Computer Vision and Pattern Recognition}}.
  \bibinfo{pages}{10975--10985}.
\newblock


\bibitem[Pavlakos et~al\mbox{.}(2024)]%
        {hamer}
\bibfield{author}{\bibinfo{person}{Georgios Pavlakos}, \bibinfo{person}{Dandan
  Shan}, \bibinfo{person}{Ilija Radosavovic}, \bibinfo{person}{Angjoo
  Kanazawa}, \bibinfo{person}{David Fouhey}, {and} \bibinfo{person}{Jitendra
  Malik}.} \bibinfo{year}{2024}\natexlab{}.
\newblock \showarticletitle{Reconstructing Hands in 3D with Transformers}. In
  \bibinfo{booktitle}{\emph{{IEEE/CVF} Conference on Computer Vision and
  Pattern Recognition, {CVPR} 2024}}. \bibinfo{pages}{9826--9836}.
\newblock


\bibitem[Peebles and Xie(2023)]%
        {peebles2023scalable}
\bibfield{author}{\bibinfo{person}{William Peebles} {and}
  \bibinfo{person}{Saining Xie}.} \bibinfo{year}{2023}\natexlab{}.
\newblock \showarticletitle{Scalable diffusion models with transformers}. In
  \bibinfo{booktitle}{\emph{Proceedings of the IEEE/CVF international
  conference on computer vision}}. \bibinfo{pages}{4195--4205}.
\newblock


\bibitem[Plappert et~al\mbox{.}(2016)]%
        {plappert2016kit}
\bibfield{author}{\bibinfo{person}{Matthias Plappert},
  \bibinfo{person}{Christian Mandery}, {and} \bibinfo{person}{Tamim Asfour}.}
  \bibinfo{year}{2016}\natexlab{}.
\newblock \showarticletitle{The kit motion-language dataset}.
\newblock \bibinfo{journal}{\emph{Big data}} \bibinfo{volume}{4},
  \bibinfo{number}{4} (\bibinfo{year}{2016}), \bibinfo{pages}{236--252}.
\newblock


\bibitem[Popov et~al\mbox{.}(2025)]%
        {popov2025camctrl3d}
\bibfield{author}{\bibinfo{person}{Stefan Popov}, \bibinfo{person}{Amit Raj},
  \bibinfo{person}{Michael Krainin}, \bibinfo{person}{Yuanzhen Li},
  \bibinfo{person}{William~T Freeman}, {and} \bibinfo{person}{Michael
  Rubinstein}.} \bibinfo{year}{2025}\natexlab{}.
\newblock \showarticletitle{CamCtrl3D: Single-Image Scene Exploration with
  Precise 3D Camera Control}.
\newblock \bibinfo{journal}{\emph{arXiv preprint arXiv:2501.06006}}
  (\bibinfo{year}{2025}).
\newblock


\bibitem[Punnakkal et~al\mbox{.}(2021)]%
        {punnakkal2021babel}
\bibfield{author}{\bibinfo{person}{Abhinanda~R Punnakkal},
  \bibinfo{person}{Arjun Chandrasekaran}, \bibinfo{person}{Nikos Athanasiou},
  \bibinfo{person}{Alejandra Quiros-Ramirez}, {and} \bibinfo{person}{Michael~J
  Black}.} \bibinfo{year}{2021}\natexlab{}.
\newblock \showarticletitle{BABEL: Bodies, action and behavior with english
  labels}. In \bibinfo{booktitle}{\emph{Proceedings of the IEEE/CVF Conference
  on Computer Vision and Pattern Recognition}}. \bibinfo{pages}{722--731}.
\newblock


\bibitem[Radford et~al\mbox{.}(2021)]%
        {radford2021learning}
\bibfield{author}{\bibinfo{person}{Alec Radford}, \bibinfo{person}{Jong~Wook
  Kim}, \bibinfo{person}{Chris Hallacy}, \bibinfo{person}{Aditya Ramesh},
  \bibinfo{person}{Gabriel Goh}, \bibinfo{person}{Sandhini Agarwal},
  \bibinfo{person}{Girish Sastry}, \bibinfo{person}{Amanda Askell},
  \bibinfo{person}{Pamela Mishkin}, \bibinfo{person}{Jack Clark},
  {et~al\mbox{.}}} \bibinfo{year}{2021}\natexlab{}.
\newblock \showarticletitle{Learning transferable visual models from natural
  language supervision}. In \bibinfo{booktitle}{\emph{International conference
  on machine learning}}. PmLR, \bibinfo{pages}{8748--8763}.
\newblock


\bibitem[Reizenstein et~al\mbox{.}(2021)]%
        {reizenstein21co3d}
\bibfield{author}{\bibinfo{person}{Jeremy Reizenstein}, \bibinfo{person}{Roman
  Shapovalov}, \bibinfo{person}{Philipp Henzler}, \bibinfo{person}{Luca
  Sbordone}, \bibinfo{person}{Patrick Labatut}, {and} \bibinfo{person}{David
  Novotny}.} \bibinfo{year}{2021}\natexlab{}.
\newblock \showarticletitle{Common Objects in 3D: Large-Scale Learning and
  Evaluation of Real-life 3D Category Reconstruction}. In
  \bibinfo{booktitle}{\emph{Proceedings of the IEEE/CVF International
  Conference on Computer Vision}}.
\newblock


\bibitem[Ren et~al\mbox{.}(2025)]%
        {ren2025gen3c}
\bibfield{author}{\bibinfo{person}{Xuanchi Ren}, \bibinfo{person}{Tianchang
  Shen}, \bibinfo{person}{Jiahui Huang}, \bibinfo{person}{Huan Ling},
  \bibinfo{person}{Yifan Lu}, \bibinfo{person}{Merlin Nimier-David},
  \bibinfo{person}{Thomas M{\"u}ller}, \bibinfo{person}{Alexander Keller},
  \bibinfo{person}{Sanja Fidler}, {and} \bibinfo{person}{Jun Gao}.}
  \bibinfo{year}{2025}\natexlab{}.
\newblock \showarticletitle{Gen3c: 3d-informed world-consistent video
  generation with precise camera control}. In
  \bibinfo{booktitle}{\emph{Proceedings of the IEEE/CVF Conference on Computer
  Vision and Pattern Recognition}}.
\newblock


\bibitem[Rombach et~al\mbox{.}(2022)]%
        {rombach2022high}
\bibfield{author}{\bibinfo{person}{Robin Rombach}, \bibinfo{person}{Andreas
  Blattmann}, \bibinfo{person}{Dominik Lorenz}, \bibinfo{person}{Patrick
  Esser}, {and} \bibinfo{person}{Bj{\"o}rn Ommer}.}
  \bibinfo{year}{2022}\natexlab{}.
\newblock \showarticletitle{High-resolution image synthesis with latent
  diffusion models}. In \bibinfo{booktitle}{\emph{Proceedings of the IEEE/CVF
  Conference on Computer Vision and Pattern Recognition}}.
  \bibinfo{pages}{10684--10695}.
\newblock


\bibitem[RunwayML(2024)]%
        {runwaygen3}
\bibfield{author}{\bibinfo{person}{RunwayML}.} \bibinfo{year}{2024}\natexlab{}.
\newblock \showarticletitle{Gen-3 Alpha}.
\newblock \bibinfo{journal}{\emph{https://runwayml.com/research/
  introducing-gen-3-alpha}} (\bibinfo{year}{2024}).
\newblock


\bibitem[Sch\"{o}nberger et~al\mbox{.}(2016)]%
        {schoenberger2016mvs}
\bibfield{author}{\bibinfo{person}{Johannes~Lutz Sch\"{o}nberger},
  \bibinfo{person}{Enliang Zheng}, \bibinfo{person}{Marc Pollefeys}, {and}
  \bibinfo{person}{Jan-Michael Frahm}.} \bibinfo{year}{2016}\natexlab{}.
\newblock \showarticletitle{Pixelwise View Selection for Unstructured
  Multi-View Stereo}. In \bibinfo{booktitle}{\emph{European conference on
  computer vision}}.
\newblock


\bibitem[Shen et~al\mbox{.}(2024)]%
        {shen2024world}
\bibfield{author}{\bibinfo{person}{Zehong Shen}, \bibinfo{person}{Huaijin Pi},
  \bibinfo{person}{Yan Xia}, \bibinfo{person}{Zhi Cen}, \bibinfo{person}{Sida
  Peng}, \bibinfo{person}{Zechen Hu}, \bibinfo{person}{Hujun Bao},
  \bibinfo{person}{Ruizhen Hu}, {and} \bibinfo{person}{Xiaowei Zhou}.}
  \bibinfo{year}{2024}\natexlab{}.
\newblock \showarticletitle{World-Grounded Human Motion Recovery via
  Gravity-View Coordinates}. In \bibinfo{booktitle}{\emph{SIGGRAPH Asia 2024
  Conference Papers}}. \bibinfo{pages}{1--11}.
\newblock


\bibitem[Sinha et~al\mbox{.}(2023)]%
        {sinha2023common}
\bibfield{author}{\bibinfo{person}{Samarth Sinha}, \bibinfo{person}{Roman
  Shapovalov}, \bibinfo{person}{Jeremy Reizenstein}, \bibinfo{person}{Ignacio
  Rocco}, \bibinfo{person}{Natalia Neverova}, \bibinfo{person}{Andrea Vedaldi},
  {and} \bibinfo{person}{David Novotny}.} \bibinfo{year}{2023}\natexlab{}.
\newblock \showarticletitle{Common pets in 3d: Dynamic new-view synthesis of
  real-life deformable categories}. In \bibinfo{booktitle}{\emph{Proceedings of
  the IEEE/CVF Conference on Computer Vision and Pattern Recognition}}.
  \bibinfo{pages}{4881--4891}.
\newblock


\bibitem[Sun et~al\mbox{.}(2024)]%
        {sun2024dimensionx}
\bibfield{author}{\bibinfo{person}{Wenqiang Sun}, \bibinfo{person}{Shuo Chen},
  \bibinfo{person}{Fangfu Liu}, \bibinfo{person}{Zilong Chen},
  \bibinfo{person}{Yueqi Duan}, \bibinfo{person}{Jun Zhang}, {and}
  \bibinfo{person}{Yikai Wang}.} \bibinfo{year}{2024}\natexlab{}.
\newblock \showarticletitle{Dimensionx: Create any 3d and 4d scenes from a
  single image with controllable video diffusion}.
\newblock \bibinfo{journal}{\emph{arXiv preprint arXiv:2411.04928}}
  (\bibinfo{year}{2024}).
\newblock


\bibitem[Tan et~al\mbox{.}(2025)]%
        {animatex}
\bibfield{author}{\bibinfo{person}{Shuai Tan}, \bibinfo{person}{Biao Gong},
  \bibinfo{person}{Xiang Wang}, \bibinfo{person}{Shiwei Zhang},
  \bibinfo{person}{Dandan Zheng}, \bibinfo{person}{Ruobing Zheng},
  \bibinfo{person}{Kecheng Zheng}, \bibinfo{person}{Jingdong Chen}, {and}
  \bibinfo{person}{Ming Yang}.} \bibinfo{year}{2025}\natexlab{}.
\newblock \showarticletitle{Animate-x: Universal character image animation with
  enhanced motion representation}. In \bibinfo{booktitle}{\emph{International
  Conference on Learning Representations}}.
\newblock


\bibitem[Tung et~al\mbox{.}(2024)]%
        {tung2024megascenes}
\bibfield{author}{\bibinfo{person}{Joseph Tung}, \bibinfo{person}{Gene Chou},
  \bibinfo{person}{Ruojin Cai}, \bibinfo{person}{Guandao Yang},
  \bibinfo{person}{Kai Zhang}, \bibinfo{person}{Gordon Wetzstein},
  \bibinfo{person}{Bharath Hariharan}, {and} \bibinfo{person}{Noah Snavely}.}
  \bibinfo{year}{2024}\natexlab{}.
\newblock \showarticletitle{Megascenes: Scene-level view synthesis at scale}.
  In \bibinfo{booktitle}{\emph{European conference on computer vision}}.
  Springer, \bibinfo{pages}{197--214}.
\newblock


\bibitem[Umeyama(1991)]%
        {umeyama1991least}
\bibfield{author}{\bibinfo{person}{Shinji Umeyama}.}
  \bibinfo{year}{1991}\natexlab{}.
\newblock \showarticletitle{Least-squares estimation of transformation
  parameters between two point patterns}.
\newblock \bibinfo{journal}{\emph{IEEE Transactions on Pattern Analysis \&
  Machine Intelligence}} \bibinfo{volume}{13}, \bibinfo{number}{04}
  (\bibinfo{year}{1991}), \bibinfo{pages}{376--380}.
\newblock


\bibitem[Veicht et~al\mbox{.}(2024)]%
        {veicht2024geocalib}
\bibfield{author}{\bibinfo{person}{Alexander Veicht},
  \bibinfo{person}{Paul-Edouard Sarlin}, \bibinfo{person}{Philipp
  Lindenberger}, {and} \bibinfo{person}{Marc Pollefeys}.}
  \bibinfo{year}{2024}\natexlab{}.
\newblock \showarticletitle{GeoCalib: Learning Single-image Calibration with
  Geometric Optimization}. In \bibinfo{booktitle}{\emph{European Conference on
  Computer Vision}}. Springer, \bibinfo{pages}{1--20}.
\newblock


\bibitem[Wang et~al\mbox{.}(2025a)]%
        {wang2025wan}
\bibfield{author}{\bibinfo{person}{Ang Wang}, \bibinfo{person}{Baole Ai},
  \bibinfo{person}{Bin Wen}, \bibinfo{person}{Chaojie Mao},
  \bibinfo{person}{Chen-Wei Xie}, \bibinfo{person}{Di Chen},
  \bibinfo{person}{Feiwu Yu}, \bibinfo{person}{Haiming Zhao},
  \bibinfo{person}{Jianxiao Yang}, \bibinfo{person}{Jianyuan Zeng},
  {et~al\mbox{.}}} \bibinfo{year}{2025}\natexlab{a}.
\newblock \showarticletitle{Wan: Open and Advanced Large-Scale Video Generative
  Models}.
\newblock \bibinfo{journal}{\emph{arXiv preprint arXiv:2503.20314}}
  (\bibinfo{year}{2025}).
\newblock


\bibitem[Wang et~al\mbox{.}(2025b)]%
        {wang2025vggt}
\bibfield{author}{\bibinfo{person}{Jianyuan Wang}, \bibinfo{person}{Minghao
  Chen}, \bibinfo{person}{Nikita Karaev}, \bibinfo{person}{Andrea Vedaldi},
  \bibinfo{person}{Christian Rupprecht}, {and} \bibinfo{person}{David
  Novotny}.} \bibinfo{year}{2025}\natexlab{b}.
\newblock \showarticletitle{VGGT: Visual Geometry Grounded Transformer}. In
  \bibinfo{booktitle}{\emph{Proceedings of the IEEE/CVF conference on computer
  vision and pattern recognition}}.
\newblock


\bibitem[Wang et~al\mbox{.}(2020)]%
        {tartanair2020iros}
\bibfield{author}{\bibinfo{person}{Wenshan Wang}, \bibinfo{person}{Delong Zhu},
  \bibinfo{person}{Xiangwei Wang}, \bibinfo{person}{Yaoyu Hu},
  \bibinfo{person}{Yuheng Qiu}, \bibinfo{person}{Chen Wang},
  \bibinfo{person}{Yafei Hu}, \bibinfo{person}{Ashish Kapoor}, {and}
  \bibinfo{person}{Sebastian Scherer}.} \bibinfo{year}{2020}\natexlab{}.
\newblock \showarticletitle{TartanAir: A Dataset to Push the Limits of Visual
  SLAM}.
\newblock  (\bibinfo{year}{2020}).
\newblock


\bibitem[Wang et~al\mbox{.}(2024a)]%
        {wang2024humanvid}
\bibfield{author}{\bibinfo{person}{Zhenzhi Wang}, \bibinfo{person}{Yixuan Li},
  \bibinfo{person}{Yanhong Zeng}, \bibinfo{person}{Youqing Fang},
  \bibinfo{person}{Yuwei Guo}, \bibinfo{person}{Wenran Liu},
  \bibinfo{person}{Jing Tan}, \bibinfo{person}{Kai Chen},
  \bibinfo{person}{Tianfan Xue}, \bibinfo{person}{Bo Dai}, {et~al\mbox{.}}}
  \bibinfo{year}{2024}\natexlab{a}.
\newblock \showarticletitle{Humanvid: Demystifying training data for
  camera-controllable human image animation}. In \bibinfo{booktitle}{\emph{The
  Thirty-eight Conference on Neural Information Processing Systems Datasets and
  Benchmarks Track}}.
\newblock


\bibitem[Wang et~al\mbox{.}(2024b)]%
        {wang2024motionctrl}
\bibfield{author}{\bibinfo{person}{Zhouxia Wang}, \bibinfo{person}{Ziyang
  Yuan}, \bibinfo{person}{Xintao Wang}, \bibinfo{person}{Yaowei Li},
  \bibinfo{person}{Tianshui Chen}, \bibinfo{person}{Menghan Xia},
  \bibinfo{person}{Ping Luo}, {and} \bibinfo{person}{Ying Shan}.}
  \bibinfo{year}{2024}\natexlab{b}.
\newblock \showarticletitle{Motionctrl: A unified and flexible motion
  controller for video generation}. In \bibinfo{booktitle}{\emph{ACM SIGGRAPH
  2024 Conference Papers}}. \bibinfo{pages}{1--11}.
\newblock


\bibitem[Xia et~al\mbox{.}(2024)]%
        {xia2024rgbd}
\bibfield{author}{\bibinfo{person}{Hongchi Xia}, \bibinfo{person}{Yang Fu},
  \bibinfo{person}{Sifei Liu}, {and} \bibinfo{person}{Xiaolong Wang}.}
  \bibinfo{year}{2024}\natexlab{}.
\newblock \showarticletitle{RGBD objects in the wild: scaling real-world 3D
  object learning from RGB-D videos}. In \bibinfo{booktitle}{\emph{Proceedings
  of the IEEE/CVF Conference on Computer Vision and Pattern Recognition}}.
  \bibinfo{pages}{22378--22389}.
\newblock


\bibitem[Xu et~al\mbox{.}(2024a)]%
        {xu2023dmv3d}
\bibfield{author}{\bibinfo{person}{Yinghao Xu}, \bibinfo{person}{Hao Tan},
  \bibinfo{person}{Fujun Luan}, \bibinfo{person}{Sai Bi}, \bibinfo{person}{Peng
  Wang}, \bibinfo{person}{Jiahao Li}, \bibinfo{person}{Zifan Shi},
  \bibinfo{person}{Kalyan Sunkavalli}, \bibinfo{person}{Gordon Wetzstein},
  \bibinfo{person}{Zexiang Xu}, {and} \bibinfo{person}{Kai Zhang}.}
  \bibinfo{year}{2024}\natexlab{a}.
\newblock \showarticletitle{DMV3D: Denoising Multi-View Diffusion using 3D
  Large Reconstruction Model}. In \bibinfo{booktitle}{\emph{International
  Conference on Learning Representations}}.
\newblock


\bibitem[Xu et~al\mbox{.}(2023)]%
        {xu2023vitpose++}
\bibfield{author}{\bibinfo{person}{Yufei Xu}, \bibinfo{person}{Jing Zhang},
  \bibinfo{person}{Qiming Zhang}, {and} \bibinfo{person}{Dacheng Tao}.}
  \bibinfo{year}{2023}\natexlab{}.
\newblock \showarticletitle{Vitpose++: Vision transformer for generic body pose
  estimation}.
\newblock \bibinfo{journal}{\emph{IEEE Transactions on Pattern Analysis and
  Machine Intelligence}} \bibinfo{volume}{46}, \bibinfo{number}{2}
  (\bibinfo{year}{2023}), \bibinfo{pages}{1212--1230}.
\newblock


\bibitem[Xu et~al\mbox{.}(2024b)]%
        {magicanimate}
\bibfield{author}{\bibinfo{person}{Zhongcong Xu}, \bibinfo{person}{Jianfeng
  Zhang}, \bibinfo{person}{Jun~Hao Liew}, \bibinfo{person}{Hanshu Yan},
  \bibinfo{person}{Jia{-}Wei Liu}, \bibinfo{person}{Chenxu Zhang},
  \bibinfo{person}{Jiashi Feng}, {and} \bibinfo{person}{Mike~Zheng Shou}.}
  \bibinfo{year}{2024}\natexlab{b}.
\newblock \showarticletitle{MagicAnimate: Temporally Consistent Human Image
  Animation using Diffusion Model}. In \bibinfo{booktitle}{\emph{{IEEE/CVF}
  Conference on Computer Vision and Pattern Recognition}}.
  \bibinfo{pages}{1481--1490}.
\newblock


\bibitem[Yang et~al\mbox{.}(2024)]%
        {yang2024direct}
\bibfield{author}{\bibinfo{person}{Shiyuan Yang}, \bibinfo{person}{Liang Hou},
  \bibinfo{person}{Haibin Huang}, \bibinfo{person}{Chongyang Ma},
  \bibinfo{person}{Pengfei Wan}, \bibinfo{person}{Di Zhang},
  \bibinfo{person}{Xiaodong Chen}, {and} \bibinfo{person}{Jing Liao}.}
  \bibinfo{year}{2024}\natexlab{}.
\newblock \showarticletitle{Direct-a-video: Customized video generation with
  user-directed camera movement and object motion}. In
  \bibinfo{booktitle}{\emph{ACM SIGGRAPH 2024 Conference Papers}}.
  \bibinfo{pages}{1--12}.
\newblock


\bibitem[Yang et~al\mbox{.}(2025)]%
        {yang2025cogvideox}
\bibfield{author}{\bibinfo{person}{Zhuoyi Yang}, \bibinfo{person}{Jiayan Teng},
  \bibinfo{person}{Wendi Zheng}, \bibinfo{person}{Ming Ding},
  \bibinfo{person}{Shiyu Huang}, \bibinfo{person}{Jiazheng Xu},
  \bibinfo{person}{Yuanming Yang}, \bibinfo{person}{Wenyi Hong},
  \bibinfo{person}{Xiaohan Zhang}, \bibinfo{person}{Guanyu Feng},
  {et~al\mbox{.}}} \bibinfo{year}{2025}\natexlab{}.
\newblock \showarticletitle{Cogvideox: Text-to-video diffusion models with an
  expert transformer}. In \bibinfo{booktitle}{\emph{International Conference on
  Learning Representations}}.
\newblock


\bibitem[You et~al\mbox{.}(2025)]%
        {you2024nvs}
\bibfield{author}{\bibinfo{person}{Meng You}, \bibinfo{person}{Zhiyu Zhu},
  \bibinfo{person}{Hui Liu}, {and} \bibinfo{person}{Junhui Hou}.}
  \bibinfo{year}{2025}\natexlab{}.
\newblock \showarticletitle{Nvs-solver: Video diffusion model as zero-shot
  novel view synthesizer}. In \bibinfo{booktitle}{\emph{International
  Conference on Learning Representations}}.
\newblock


\bibitem[Yu et~al\mbox{.}(2024)]%
        {yu2024viewcrafter}
\bibfield{author}{\bibinfo{person}{Wangbo Yu}, \bibinfo{person}{Jinbo Xing},
  \bibinfo{person}{Li Yuan}, \bibinfo{person}{Wenbo Hu},
  \bibinfo{person}{Xiaoyu Li}, \bibinfo{person}{Zhipeng Huang},
  \bibinfo{person}{Xiangjun Gao}, \bibinfo{person}{Tien-Tsin Wong},
  \bibinfo{person}{Ying Shan}, {and} \bibinfo{person}{Yonghong Tian}.}
  \bibinfo{year}{2024}\natexlab{}.
\newblock \showarticletitle{Viewcrafter: Taming video diffusion models for
  high-fidelity novel view synthesis}.
\newblock \bibinfo{journal}{\emph{arXiv preprint arXiv:2409.02048}}
  (\bibinfo{year}{2024}).
\newblock


\bibitem[Zhang et~al\mbox{.}(2023)]%
        {zhang2023adding}
\bibfield{author}{\bibinfo{person}{Lvmin Zhang}, \bibinfo{person}{Anyi Rao},
  {and} \bibinfo{person}{Maneesh Agrawala}.} \bibinfo{year}{2023}\natexlab{}.
\newblock \showarticletitle{Adding conditional control to text-to-image
  diffusion models}. In \bibinfo{booktitle}{\emph{Proceedings of the IEEE/CVF
  international conference on computer vision}}. \bibinfo{pages}{3836--3847}.
\newblock


\bibitem[Zhao et~al\mbox{.}(2022)]%
        {zhao2022particlesfm}
\bibfield{author}{\bibinfo{person}{Wang Zhao}, \bibinfo{person}{Shaohui Liu},
  \bibinfo{person}{Hengkai Guo}, \bibinfo{person}{Wenping Wang}, {and}
  \bibinfo{person}{Yong-Jin Liu}.} \bibinfo{year}{2022}\natexlab{}.
\newblock \showarticletitle{ParticleSfM: Exploiting Dense Point Trajectories
  for Localizing Moving Cameras in the Wild}. In
  \bibinfo{booktitle}{\emph{European conference on computer vision (ECCV)}}.
\newblock


\bibitem[Zheng et~al\mbox{.}(2025a)]%
        {zheng2024cami2v}
\bibfield{author}{\bibinfo{person}{Guangcong Zheng}, \bibinfo{person}{Teng Li},
  \bibinfo{person}{Rui Jiang}, \bibinfo{person}{Yehao Lu}, \bibinfo{person}{Tao
  Wu}, {and} \bibinfo{person}{Xi Li}.} \bibinfo{year}{2025}\natexlab{a}.
\newblock \showarticletitle{Cami2v: Camera-controlled image-to-video diffusion
  model}. In \bibinfo{booktitle}{\emph{International Conference on Learning
  Representations}}.
\newblock


\bibitem[Zheng et~al\mbox{.}(2025b)]%
        {zheng2025vidcraft3}
\bibfield{author}{\bibinfo{person}{Sixiao Zheng}, \bibinfo{person}{Zimian
  Peng}, \bibinfo{person}{Yanpeng Zhou}, \bibinfo{person}{Yi Zhu},
  \bibinfo{person}{Hang Xu}, \bibinfo{person}{Xiangru Huang}, {and}
  \bibinfo{person}{Yanwei Fu}.} \bibinfo{year}{2025}\natexlab{b}.
\newblock \showarticletitle{VidCRAFT3: Camera, Object, and Lighting Control for
  Image-to-Video Generation}.
\newblock \bibinfo{journal}{\emph{arXiv preprint arXiv:2502.07531}}
  (\bibinfo{year}{2025}).
\newblock


\bibitem[Zhou et~al\mbox{.}(2024)]%
        {realisdance}
\bibfield{author}{\bibinfo{person}{Jingkai Zhou}, \bibinfo{person}{Benzhi
  Wang}, \bibinfo{person}{Weihua Chen}, \bibinfo{person}{Jingqi Bai},
  \bibinfo{person}{Dongyang Li}, \bibinfo{person}{Aixi Zhang},
  \bibinfo{person}{Hao Xu}, \bibinfo{person}{Mingyang Yang}, {and}
  \bibinfo{person}{Fan Wang}.} \bibinfo{year}{2024}\natexlab{}.
\newblock \showarticletitle{RealisDance: Equip controllable character animation
  with realistic hands}.
\newblock \bibinfo{journal}{\emph{arXiv preprint arXiv:2409.06202}}
  (\bibinfo{year}{2024}).
\newblock


\bibitem[Zhou et~al\mbox{.}(2025b)]%
        {zhou2025RealisDance}
\bibfield{author}{\bibinfo{person}{Jingkai Zhou}, \bibinfo{person}{Yifan Wu},
  \bibinfo{person}{Shikai Li}, \bibinfo{person}{Min Wei}, \bibinfo{person}{Chao
  Fan}, \bibinfo{person}{Weihua Chen}, \bibinfo{person}{Wei Jiang}, {and}
  \bibinfo{person}{Fan Wang}.} \bibinfo{year}{2025}\natexlab{b}.
\newblock \showarticletitle{RealisDance-DiT: Simple yet Strong Baseline towards
  Controllable Character Animation in the Wild}.
\newblock \bibinfo{journal}{\emph{arXiv preprint arXiv:2504.14977}}
  (\bibinfo{year}{2025}).
\newblock


\bibitem[Zhou et~al\mbox{.}(2025a)]%
        {zhou2025stable}
\bibfield{author}{\bibinfo{person}{Jensen~Jinghao Zhou}, \bibinfo{person}{Hang
  Gao}, \bibinfo{person}{Vikram Voleti}, \bibinfo{person}{Aaryaman Vasishta},
  \bibinfo{person}{Chun-Han Yao}, \bibinfo{person}{Mark Boss},
  \bibinfo{person}{Philip Torr}, \bibinfo{person}{Christian Rupprecht}, {and}
  \bibinfo{person}{Varun Jampani}.} \bibinfo{year}{2025}\natexlab{a}.
\newblock \showarticletitle{STABLE VIRTUAL CAMERA: Generative View Synthesis
  with Diffusion Models}.
\newblock \bibinfo{journal}{\emph{arXiv e-prints}} (\bibinfo{year}{2025}),
  \bibinfo{pages}{arXiv--2503}.
\newblock


\bibitem[Zhou et~al\mbox{.}(2018)]%
        {zhou2018Stereo}
\bibfield{author}{\bibinfo{person}{Tinghui Zhou}, \bibinfo{person}{Richard
  Tucker}, \bibinfo{person}{John Flynn}, \bibinfo{person}{Graham Fyffe}, {and}
  \bibinfo{person}{Noah Snavely}.} \bibinfo{year}{2018}\natexlab{}.
\newblock \showarticletitle{Stereo magnification: learning view synthesis using
  multiplane images}.
\newblock \bibinfo{journal}{\emph{ACM Trans. Graph.}} \bibinfo{volume}{37},
  \bibinfo{number}{4}, Article \bibinfo{articleno}{65} (\bibinfo{date}{July}
  \bibinfo{year}{2018}), \bibinfo{numpages}{12}~pages.
\newblock
\showISSN{0730-0301}
\href{https://doi.org/10.1145/3197517.3201323}{doi:\nolinkurl{10.1145/3197517.3201323}}


\bibitem[Zhu et~al\mbox{.}(2024)]%
        {champ}
\bibfield{author}{\bibinfo{person}{Shenhao Zhu}, \bibinfo{person}{Junming~Leo
  Chen}, \bibinfo{person}{Zuozhuo Dai}, \bibinfo{person}{Zilong Dong},
  \bibinfo{person}{Yinghui Xu}, \bibinfo{person}{Xun Cao}, \bibinfo{person}{Yao
  Yao}, \bibinfo{person}{Hao Zhu}, {and} \bibinfo{person}{Siyu Zhu}.}
  \bibinfo{year}{2024}\natexlab{}.
\newblock \showarticletitle{Champ: Controllable and Consistent Human Image
  Animation with 3D Parametric Guidance}. In \bibinfo{booktitle}{\emph{European
  Conference on Computer Vision}}, Vol.~\bibinfo{volume}{15113}.
  \bibinfo{pages}{145--162}.
\newblock


\end{thebibliography}

\appendix

\section{Border Impact}

This study delves into the realm of controllable video generation. The remarkable generative capabilities of AI-generated content (AIGC) can inadvertently lead to the creation of misleading information or fabricated visuals. Consequently, we sincerely urge users to remain vigilant regarding these issues.
Moreover, issues of privacy and consent must be taken into account, as generative models are always developed using extensive datasets. It's also crucial to acknowledge that such models can potentially reinforce biases in the training data, which may result in unjust outcomes. Therefore, we advise users to be responsible and inclusive when utilizing these generative models.
It is important to mention that our method concentrates solely on technical aspects. All pre-trained models and training videos referenced in this study are publicly available.

\section{Details of Uni3C}

\begin{figure}
\centering
\includegraphics[width=0.8\linewidth]{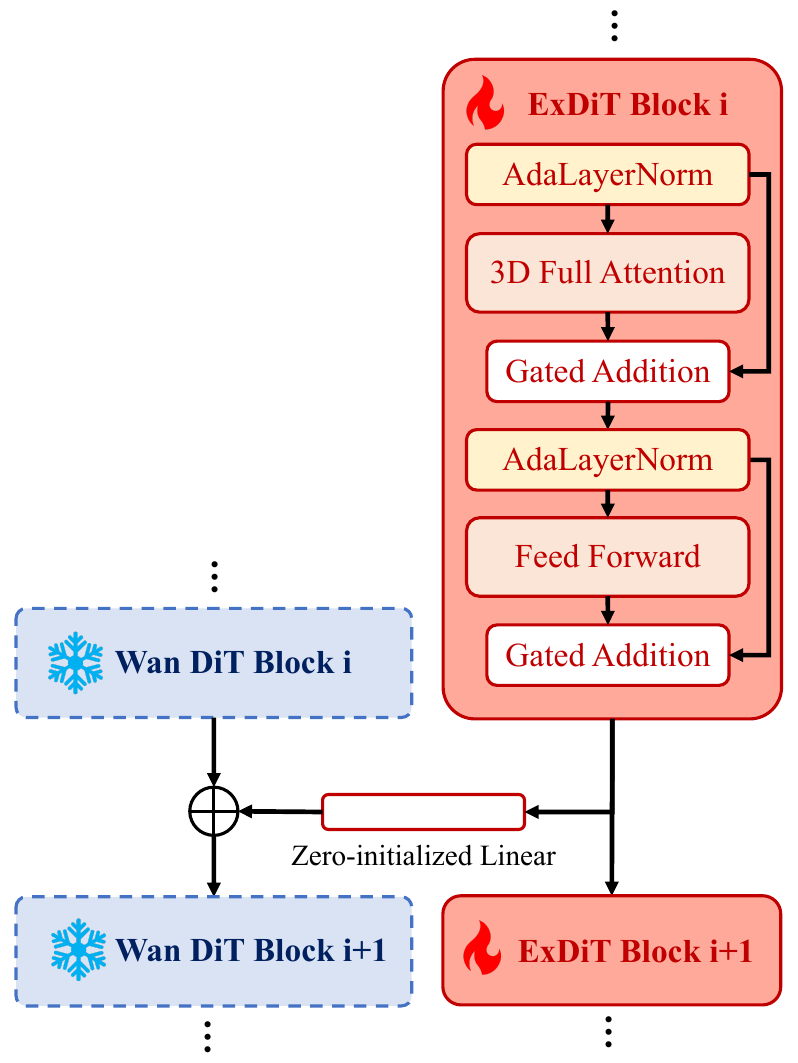}
\vspace{-0.1in}
   \caption{\textbf{Detailed model architecture of PCDController.}
    \label{fig:arch_detail}}
\vspace{-0.1in}
\end{figure}

\subsection{Architecture Details of PCDController}
As mentioned in the main paper, our PCDController is built upon a simplified DiT module. We apply detailed ablation studies and demonstrate that the 20-layer external DiT branch with 1024 hidden size enjoys a good balance between both controllability and generalization. 
Detailed architecture of PCDController is illustrated in \Cref{fig:arch_detail}. 
We follow the DiT design in CogVideoX~\cite{yang2025cogvideox}, including AdaLayerNorm, 3D attention, and Feed Forward Network (FFN), without the textual branch.
The output features are added to the main branch of Wan-I2V block by block via zero-initialized projection layers.

\begin{figure}
\centering
\includegraphics[width=0.75\linewidth]{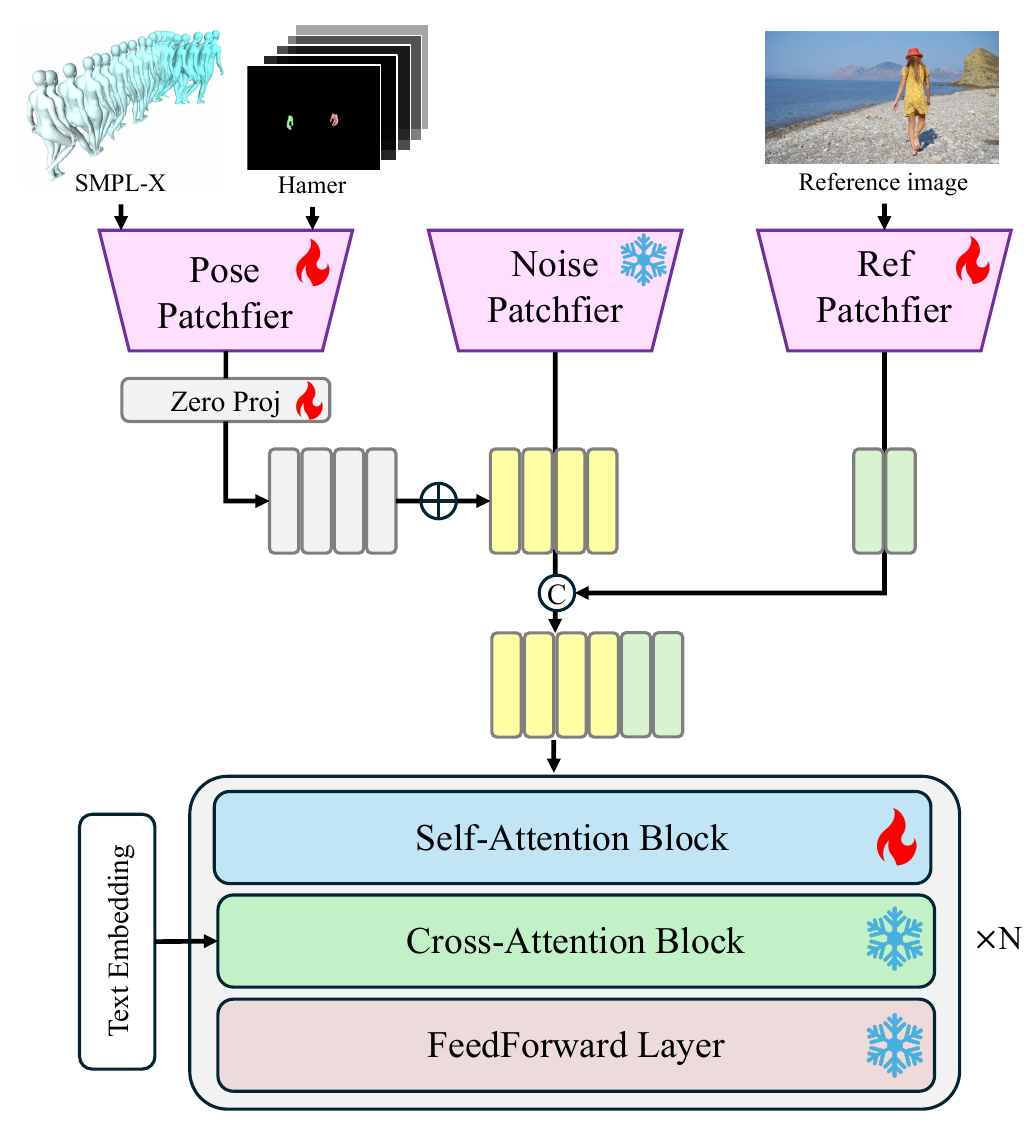}
\vspace{-0.1in}
   \caption{\revise{\textbf{Detailed architecture of Realisdance-DiT~\cite{zhou2025RealisDance}.}}
    \label{fig:realisdance_dit_detail}}
\vspace{-0.1in}
\end{figure}

\subsection{Architecture Details of Realisdance-DiT}
\revise{We show the model architecture of Realisdance-DiT~\cite{zhou2025RealisDance} in \Cref{fig:realisdance_dit_detail} employed for our unified controls for both camera and human poses.
Formally, Realisdance-DiT is fine-tuned from the pre-trained Wan-I2V, while features of human condition signals (SMPL-X and Hamer) are added to the inputs, and the reference image feature is spatially concatenated to the inputs to confirm fine-grained details. During the training of Realisdance-DiT, only convolution-based pose and reference patchfiers and self-attention layers of DiT are trainable.}

\subsection{High-Resolution Inference}

While PCDController is trained under 480p multi-view images ($480\times768$, $512\times720$, $608\times608$, $720\times512$, $768\times480$), we find that this model also performs well under 720p video generation ($720\times1280$, $800\times1152$, $960\times960$, $1152\times800$, $1280\times720$) without specific fine-tuning, as empirically verified in our qualitative results in the supplementary.
Specifically, for the inference of 720p, we first render the conditional videos of warped point clouds with PyTorch3D under 480p. Then we replace the first frame with a 720p reference image as the high-resolution guidance. Thus, PCDController can produce high-quality 720p videos extended from the high-resolution reference, while the low-resolution point cloud renderings are only considered as auxiliary camera signals.

\subsection{Inference Efficiency}

\begin{table}
\centering
\small
\caption{\textbf{Inference efficiency of Uni3C.} Our model only increases a little inference time compared to the baseline method.\label{tab:efficiency}}
\vspace{-0.1in}
\begin{tabular}{l|cc}
\toprule 
\multirow{2}{*}{Methods} & \multicolumn{2}{c}{Inference Times}\tabularnewline
 & 480$\times$768 & 720$\times$1280\tabularnewline
\midrule
RealisDance-DiT & 50.5s & 182.7s\tabularnewline
Uni3C & 62.3s (+23\%) & 213.4s (+17\%)\tabularnewline
\bottomrule
\end{tabular}
\vspace{-0.1in}
\end{table}

\begin{table}
\centering
\small
\caption{\revise{\textbf{Efficiency of other components of Uni3C.}\label{tab:efficiency2}}}
\vspace{-0.1in}
\begin{tabular}{ccc}
\toprule 
PyTorch3D-Rendering & 2D-Keypoint+Alignment & GeoCalib \tabularnewline
\midrule
4.341s & 0.577s & 1.459s\tabularnewline
\bottomrule
\end{tabular}
\vspace{-0.1in}
\end{table}

We evaluate the inference efficiency of Uni3C in comparison to the baseline model, RealisDance-DiT~\cite{zhou2025RealisDance}, across two different resolutions, as detailed in \Cref{tab:efficiency}.
Note that the efficiency of RealisDance-DiT~\cite{zhou2025RealisDance} is almost the same as the basic VDM, Wan-I2V~\cite{wang2025wan}, while only a few convolution layers are newly incorporated to encode additional conditions.
For this analysis, we utilize the official settings from Wan2.1~\cite{wang2025wan}, which include a 40-step denoising process and a classifier-free guidance scale of 5. The inference environment is configured using a sequence parallel based on 8 Nvidia H100 GPUs.
\revise{We further provide other detailed time costs of Uni3C in \Cref{tab:efficiency2}.}
Overall, Uni3C demonstrates impressive efficiency in this setup.

\section{Details of Datasets}

\begin{table}
\centering
\caption{\textbf{Dataset details of training PCDController.} 
The training datasets include Co3Dv2~\cite{reizenstein21co3d}, DL3DV~\cite{ling2024dl3dv}, RE10K~\cite{zhou2018Stereo}, ACID~\cite{liu2021infinite}, Tartainair~\cite{tartanair2020iros}, Map-Free-Reloc~\cite{arnold2022map}, WildRGBD~\cite{xia2024rgbd}, COP3D~\cite{sinha2023common}, and UCo3D~\cite{liu2025uncommon}.
We dynamically sample subsets for each dataset across training epochs.\label{tab:dataset_details}}
\vspace{-0.1in}
{\footnotesize{}}%
\setlength{\tabcolsep}{2.2pt} 
\begin{tabular}{lccccccc}
\toprule 
 & \multirow{2}{*}{{\footnotesize{}Indoor}} & \multirow{2}{*}{{\footnotesize{}Outdoor}} & \multirow{2}{*}{{\footnotesize{}Object}} & \multirow{2}{*}{{\footnotesize{}Synthetic}} & \multicolumn{3}{c}{{\footnotesize{}Scene Number}}\tabularnewline
 &  &  &  &  & {\footnotesize{}Train} & {\footnotesize{}Validation} & {\footnotesize{}Epoch}\tabularnewline
\midrule 
{\footnotesize{}Co3Dv2} & {\footnotesize{}$\checkmark$} & {\footnotesize{}$\checkmark$} & {\footnotesize{}$\checkmark$} &  & {\footnotesize{}24,437} & {\footnotesize{}53} & {\footnotesize{}2,252}\tabularnewline
{\footnotesize{}DL3DV} & {\footnotesize{}$\checkmark$} & {\footnotesize{}$\checkmark$} &  &  & {\footnotesize{}9,808} & {\footnotesize{}250} & {\footnotesize{}19,868}\tabularnewline
{\footnotesize{}RE10K} & {\footnotesize{}$\checkmark$} & {\footnotesize{}$\checkmark$} &  &  & {\footnotesize{}20,259} & {\footnotesize{}50} & {\footnotesize{}7,528}\tabularnewline
{\footnotesize{}ACID} &  & {\footnotesize{}$\checkmark$} &  &  & {\footnotesize{}2,575} & {\footnotesize{}20} & {\footnotesize{}1,047}\tabularnewline
{\footnotesize{}Tartainair} & {\footnotesize{}$\checkmark$} & {\footnotesize{}$\checkmark$} &  & {\footnotesize{}$\checkmark$} & {\footnotesize{}2,834} & {\footnotesize{}50} & {\footnotesize{}2,834}\tabularnewline
{\footnotesize{}Map-Free-Reloc} & {\footnotesize{}$\checkmark$} & {\footnotesize{}$\checkmark$} &  &  & {\footnotesize{}892} & {\footnotesize{}18} & {\footnotesize{}1,784}\tabularnewline
{\footnotesize{}WildRGBD} & {\footnotesize{}$\checkmark$} & {\footnotesize{}$\checkmark$} & {\footnotesize{}$\checkmark$} &  & {\footnotesize{}22,105} & {\footnotesize{}46} & {\footnotesize{}2,300}\tabularnewline
{\footnotesize{}COP3D} & {\footnotesize{}$\checkmark$} & {\footnotesize{}$\checkmark$} &  &  & {\footnotesize{}2,109} & {\footnotesize{}20} & {\footnotesize{}2,109}\tabularnewline
{\footnotesize{}UCo3D} & {\footnotesize{}$\checkmark$} & {\footnotesize{}$\checkmark$} & {\footnotesize{}$\checkmark$} &  & {\footnotesize{}161,591} & {\footnotesize{}99} & {\footnotesize{}10,000}\tabularnewline
\bottomrule 
\end{tabular}{\footnotesize\par}
\end{table}

We summarized our training data setting in \Cref{tab:dataset_details}. 
A sampling strategy across epochs ensures balanced sample scales over different data domains. 
Formally, we pay more attention to the data with high-quality images and complex camera trajectories of real-world scenarios like DL3DV~\cite{ling2024dl3dv} and UCo3D~\cite{liu2025uncommon}.

\begin{figure*}[htb!]
\centering
\includegraphics[width=1.0\linewidth]{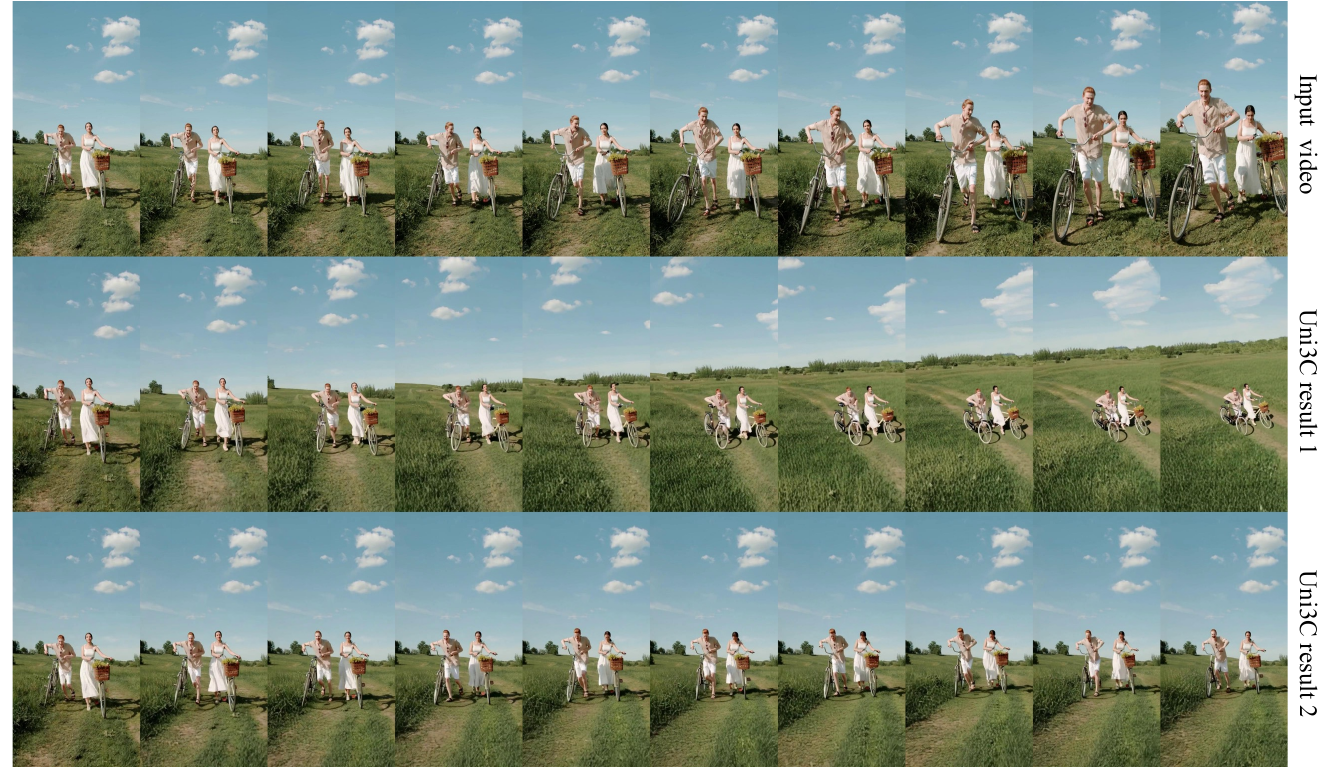}
\vspace{-0.15in}
   \caption{\revise{\textbf{Results of Uni3C under the two-person scenario.} We operate two different camera trajectories, while their human motions follow the input video. \label{fig:two_person}}}
\vspace{-0.1in}
\end{figure*}

\section{Inference Details of Camera Control}

During the inference phase, we begin by extracting the monocular depth of the reference view using Depth-Pro~\cite{Bochkovskii2024depthpro}. We then establish the foreground depth medium, defined as the rotation radius, to determine the placement of the initial camera.
For extracting the foreground mask, we utilize CarveKit~\footnote{\url{https://github.com/OPHoperHPO/image-background-remove-tool}}.
In cases where no foreground is detected, the entire image is treated as the foreground.
Our camera control is based on two primitives:
\begin{itemize}
    \item \textbf{Rotation.} We define the rotation along azimuth and elevation, respectively. The rotation radius is used to control the distance to the foreground, serving as the center of rotation. 
    \item \textbf{Translation:} We implement translation along the x, y, and z axes. The translation values are constrained within the range of $[0, 1.0]$, which are then multiplied by the rotation radius to ensure that translations do not become excessively large or small.
\end{itemize}
Overall, the camera control mechanism proposed in this paper is flexible and effective enough to handle most downstream tasks.



\section{Multiple-Person Discussion}

\revise{Uni3C builds upon RealisDance-DiT~\cite{zhou2025RealisDance}, which is explicitly tailored for single-person scenarios. Notably, we observe that RealisDance-DiT retains functionality even in two-person scenarios, as verified by the case presented in~\cite{zhou2025RealisDance} and visualized in \Cref{fig:two_person}.
Formally, we separately align two persons' SMPL-X representations, guided by their respective 3D keypoints.
However, injecting additional control signals from more people would hinder the performance. This outcome is unsurprising, as multi-person control is significantly outside the scope of RealisDance-DiT training.
As a plug-and-play module, Uni3C can be integrated with other backbones that are natively designed for addressing multi-person control in the future.}

\end{document}